\documentclass[conference]{IEEEtran}
\usepackage{times}

\usepackage[numbers]{natbib}
\usepackage{multicol}
\usepackage[bookmarks=true]{hyperref}

\let\labelindent\relax

\usepackage[numbers]{natbib}
\usepackage{multicol}
\usepackage[bookmarks=true]{hyperref}
\usepackage{amsmath}
\usepackage{amssymb}
\usepackage{color}
\usepackage{graphicx}
\usepackage{appendix}
\graphicspath{{images/}}
\usepackage{amsmath}

\usepackage{xspace}

\usepackage{setspace}
\usepackage{amsmath}
\usepackage{amsfonts}
\usepackage{amssymb}
\usepackage{amsthm}
\usepackage{bm}
\usepackage{bbm}
\usepackage{mathtools}
\usepackage{enumitem}
\usepackage{thmtools,thm-restate}
\usepackage{algorithm}
\usepackage{algorithmic}
\usepackage{color}
\usepackage{graphicx}
\usepackage{comment}
\def\Rbb{\mathbb{R}}

\def\R{\Rbb}

\def\*{\star}

\newcommand{\abs}[1]{ \left| #1 \right|  }
\newcommand{\tr}[1]{ \mathrm{tr}\left( #1\right)}

\usepackage[dvipsnames]{xcolor}

\renewcommand{\tr}{{T}}

\newcommand{\paral}{{\!/\mkern-5mu/\!}}

\newcommand{\Lag}{\mathcal{L}}
\newcommand{\Ham}{\mathcal{H}}

\newcommand{\p}{\mathbf{p}}
\newcommand{\q}{\mathbf{q}}
\newcommand{\qd}{{\dot{\q}}}
\newcommand{\qdd}{{\ddot{\q}}}

\newcommand{\vv}{\mathbf{v}}

\newcommand{\x}{\mathbf{x}}
\newcommand{\xd}{{\dot{\x}}}
\newcommand{\xdd}{{\ddot{\x}}}
\newcommand{\y}{\mathbf{y}}
\newcommand{\yd}{{\dot{\y}}}

\newcommand{\f}{\mathbf{f}}
\newcommand{\h}{\mathbf{h}}

\newcommand{\zero}{\mathbf{0}}

\newcommand{\J}{\mathbf{J}}
\newcommand{\Jd}{{\dot{\J}}}

\newcommand{\B}{\mathbf{B}}

\newcommand{\G}{\mathbf{G}}

\newcommand{\I}{\mathbf{I}}

\newcommand{\M}{\mathbf{M}}

\newcommand{\mP}{\mathbf{P}}
\newcommand{\mR}{\mathbf{R}}

\newcommand{\wt}[1]{{\widetilde{#1}}}

\usepackage{amsmath}  % Conflicts w/ WAFR style template
\usepackage{amssymb}
\usepackage{accents}  % Conflicts w/ WAFR style template
\usepackage{amsthm}

\theoremstyle{plain}
\newtheorem{theorem}{Theorem}[section]

\newtheorem{proposition}[theorem]{Proposition}

\theoremstyle{definition}

\theoremstyle{remark}

\makeatletter
\let\save@mathaccent\mathaccent
\newcommand*\if@single[3]{%
  \setbox0\hbox{${\mathaccent"0362{#1}}^H$}%
  \setbox2\hbox{${\mathaccent"0362{\kern0pt#1}}^H$}%
  \ifdim\ht0=\ht2 #3\else #2\fi
  }
\newcommand*\rel@kern[1]{\kern#1\dimexpr\macc@kerna}
\newcommand*\widebar[1]{\@ifnextchar^{{\wide@bar{#1}{0}}}{\wide@bar{#1}{1}}}
\newcommand*\wide@bar[2]{\if@single{#1}{\wide@bar@{#1}{#2}{1}}{\wide@bar@{#1}{#2}{2}}}
\newcommand*\wide@bar@[3]{%
  \begingroup
  \def\mathaccent##1##2{%
    \let\mathaccent\save@mathaccent
    \if#32 \let\macc@nucleus\first@char \fi
    \setbox\z@\hbox{$\macc@style{\macc@nucleus}_{}$}%
    \setbox\tw@\hbox{$\macc@style{\macc@nucleus}{}_{}$}%
    \dimen@\wd\tw@
    \advance\dimen@-\wd\z@
    \divide\dimen@ 3
    \@tempdima\wd\tw@
    \advance\@tempdima-\scriptspace
    \divide\@tempdima 10
    \advance\dimen@-\@tempdima
    \ifdim\dimen@>\z@ \dimen@0pt\fi
    \rel@kern{0.6}\kern-\dimen@
    \if#31
      \overline{\rel@kern{-0.6}\kern\dimen@\macc@nucleus\rel@kern{0.4}\kern\dimen@}%
      \advance\dimen@0.4\dimexpr\macc@kerna
      \let\final@kern#2%
      \ifdim\dimen@<\z@ \let\final@kern1\fi
      \if\final@kern1 \kern-\dimen@\fi
    \else
      \overline{\rel@kern{-0.6}\kern\dimen@#1}%
    \fi
  }%
  \macc@depth\@ne
  \let\math@bgroup\@empty \let\math@egroup\macc@set@skewchar
  \mathsurround\z@ \frozen@everymath{\mathgroup\macc@group\relax}%
  \macc@set@skewchar\relax
  \let\mathaccentV\macc@nested@a
  \if#31
    \macc@nested@a\relax111{#1}%
  \else
    \def\gobble@till@marker##1\endmarker{}%
    \futurelet\first@char\gobble@till@marker#1\endmarker
    \ifcat\noexpand\first@char A\else
      \def\first@char{}%
    \fi
    \macc@nested@a\relax111{\first@char}%
  \fi
  \endgroup
}
\makeatother
\newcommand{\mandy}[1]{\noindent\textcolor{DarkOrchid}{Mandy:} \textcolor{CadetBlue}{#1}}

\newcommand{\nathan}[1]{\noindent\textcolor{ForestGreen}{Nathan:} \textcolor{MidnightBlue}{#1}}

\newcommand{\fullversiononly}{\iffalse}
\newcommand{\compressedversiononly}{\iftrue}

\renewcommand{\baselinestretch}{.99}

\title{Geometric Fabrics for the Acceleration-based Design of Robotic Motion}

\author{\authorblockN{
Mandy Xie\authorrefmark{1}\authorrefmark{2}, 
Karl Van Wyk\authorrefmark{1},
Anqi Li\authorrefmark{1}\authorrefmark{3} and
Muhammad Asif Rana\authorrefmark{2},\\
Qian Wan\authorrefmark{1},
Dieter Fox\authorrefmark{1}\authorrefmark{3},
Byron Boots\authorrefmark{1}\authorrefmark{3},
Nathan D. Ratliff\authorrefmark{1}
}
\authorblockA{
\authorrefmark{1}NVIDIA (\{nratliff,kvanwyk\}@nvidia.com);
\authorrefmark{2}Georgia Tech;
\authorrefmark{3}University of Washington}
}

\begin{document}
\makeatletter
\let\@oldmaketitle\@maketitle% Store \@maketitle
\renewcommand{\@maketitle}{\@oldmaketitle% Update \@maketitle to insert...
  \includegraphics[width=0.999\linewidth]
    {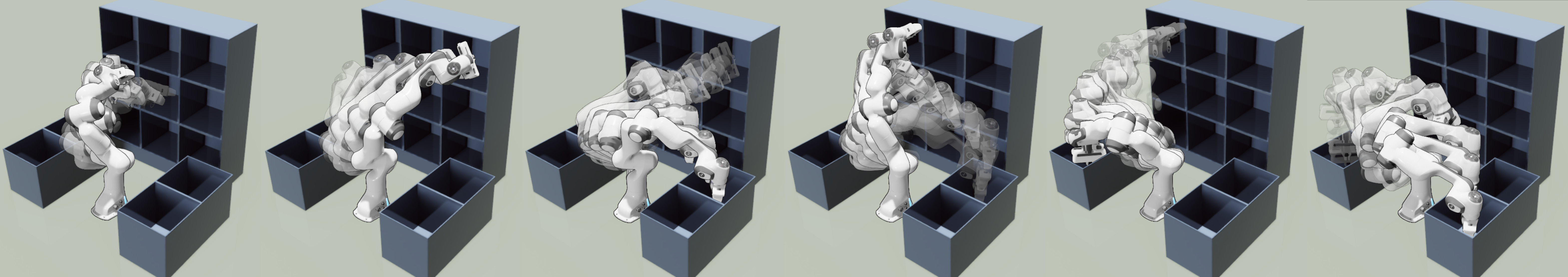} \\%[0.25em]
  \refstepcounter{figure}\footnotesize{Fig.~\thefigure. Robot seamlessly navigating cubbies using a geometric fabric that enables obstacle and joint limit avoidance, redundancy resolution, and goal reaching.}
  \label{fig:CubbyResults} \medskip \vspace{-15pt}}% ... an image
\makeatother

\maketitle
\thispagestyle{empty}
\pagestyle{empty}

%\author{\authorblockN{Michael Shell}
%\authorblockA{School of Electrical and\\Computer Engineering\\
%Georgia Institute of Technology\\
%Atlanta, Georgia 30332--0250\\
%Email: mshell@ece.gatech.edu}
%\and
%\authorblockA{Twentieth Century Fox\\
%\and
%\authorblockN{James Kirk\\ and Montgomery Scott}
%\authorblockA{Starfleet Academy\\
%San Francisco, California 96678-2391\\

% avoiding spaces at the end of the author lines is not a problem with
% conference papers because we don't use \thanks or \IEEEmembership

%Homer Simpson\authorrefmark{2},
%Montgomery Scott\authorrefmark{3} and
%Eldon Tyrell\authorrefmark{4}}
%Telephone: (800) 555--1212, Fax: (888) 555--1212}
%\authorblockA{\authorrefmark{4}Tyrell Inc., 123 Replicant Street, Los Angeles, California 90210--4321}}

\maketitle

\begin{abstract}
This paper describes the pragmatic design and construction of geometric fabrics for shaping a robot's task-independent nominal behavior, capturing behavioral components such as obstacle avoidance, joint limit avoidance, redundancy resolution, global navigation heuristics, etc. Geometric fabrics constitute the most concrete incarnation of a new mathematical formulation for reactive behavior called optimization fabrics. Fabrics generalize recent work on Riemannian Motion Policies (RMPs); they add provable stability guarantees and improve design consistency while promoting the intuitive acceleration-based principles of modular design that make RMPs successful. We describe a suite of mathematical modeling tools that practitioners can employ in practice and demonstrate both how to mitigate system complexity by constructing behaviors layer-wise and how to employ these tools to design robust, strongly-generalizing, policies that solve practical problems one would expect to find in industry applications. Our system exhibits intelligent global navigation behaviors expressed entirely as provably stable fabrics with zero planning or state machine governance.
\end{abstract}

\IEEEpeerreviewmaketitle

\section{INTRODUCTION}

%\fullversiononly
Motion generation is one of the fundamental problems of robotics. 
%\fi
Fast, reactive motion is essential for most modern tasks, especially in highly-dynamic and uncertain collaborative environments. We describe here a set of tools built from {\em geometric fabrics}, derived from our recent theory of {\em optimization fabrics} (see Section III of supplemental paper \cite{ratliffoptimizationfabricsupp}),
for the direct construction of {\em stable} robotic behavior in modular parts.\footnote{The term {\em fabric} is used analogously to the term {\em fabric of spacetime} from theoretical physics, but formalizes the idea as a second-order nonlinear differential equation characterizing an {\em unbiased} nominal behavior.} 
Fabrics define a nominal behavior independent of a specific task, capturing cross-task commonalities like joint-limit avoidance, obstacle avoidance, and redundancy resolution, and implement a task as an optimization problem across the fabric. The fabric defines behavior by shaping the optimization path. 
% \compressedversiononly
% % Entirely removed, no paragraph break either
% \else % Full version

For instance, the goal of a given task may be to reach a target point with the robot's end-effector. En route, the system should avoid obstacles and joint limits, resolve redundancy intelligently, implement global navigation heuristics, and may even shape the end-effector path to approach the target from a specific direction. The task is the optimization problem characterizing the end-effector's target as its local minimum; the fabric captures everything else about the behavior we want the robot to exhibit as it optimizes that objective.

The theory of fabrics was motivated by the empirical success of Riemannian Motion Policies (RMPs) \cite{ratliff2018rmps,cheng2018rmpflow} which have been shown to demonstrate flexible robust performance on real-world reactive and adaptive tasks. RMPs are intuitive. Designers can build behaviors as acceleration policies on different spaces and provide velocity dependent weight matrices defining how they should combine as metric weighted averages. Unfortunately, they have no theoretical guarantees of stability and empirically some care is needed to tune them well so the contributing RMP terms don't conflict with one another. Fabrics, on the other hand, are fundamentally {\em unbiased} in a rigorous sense (see Section III of supplemental paper \cite{ratliffoptimizationfabricsupp}), meaning practically that the underlying fabric does not prevent the system from achieving task goals. And they fundamentally engender asymptotic stability making them an alluring framework for behavioral design.

Geometric fabrics are a special type of fabric that expresses its unbiased nominal behavior as a generalized nonlinear geometry in the robot's configuration space; they constitute the most concrete incarnation of optimization fabric and capture many of intuitive properties that make RMPs so powerful, such as acceleration-based policy design and independent priority metric specification. Similar to RMPs, geometric fabrics can be conveniently constructed in parts distributed across a transform tree of relevant task spaces. Importantly, they inherit key theoretical properties from the theory of fabrics, including stability and their unbiased behavior. Additionally, due to their construction as nonlinear geometries of paths, geometric fabrics exhibit a characteristic geometric consistency which allows us both to construct them layer-wise to mitigate design complexity and to independently control execution speed by accelerate along the direction of motion without affecting the overall the behavior. Geometric fabrics capture RMP intuition but with important gains from their theoretical foundation.

% \fi % End full version only
% \compressedversiononly % This will connect to the paragraph above the preceeding section
% Geometric fabrics specifically characterize behavior as generalized nonlinear geometries. We derive tools for designing these fabrics and demonstrate their utility on a Franka Panda robot in a standard setting commonplace in industry.
% \else % Full version

We will show that a wide range of robotic behavior can be captured purely by its geometric fabric. We detail a pragmatic collection of modeling tools derived from this framework, and present experimental results on a Franka Panda robotic manipulator fluently navigating furniture mimicking problems common in logistics or industrial settings.
% \fi % End compressed version only

\compressedversiononly
%\karl{I commented out below (check tex file) discussion on limitations of previous works (GDS, OSC, etc.) since we aren't actually comparing with them here anymore. Maybe we can briefly just mention these issues in related work?}
%We experimentally verify in Section~\ref{sec:PlanarExperiments} a theoretical result discussed in \cite{optimizationFabricsForBehavioralDesignArXiv2020} (Section I.A) that shows a fundamental limitation of many existing techniques, such as operational space control \ref{KhatibOperationalSpaceControl1987,Peters_OpSpaceControl_AR_2008}, geometric control \cite{bullo2004geometric}, and geometric dynamical systems \cite{cheng2018rmpflow}, which we know as Lagrangian fabrics within the theory of optimization fabrics. Essentially, we see that all of those systems {\em either} have fundamentally limited expressivity {\em or} they must express behavior through objective potentials which renders them prone to conflict.
\else
Importantly, geometric fabrics are a class of stable RMPs.
In \cite{optimizationFabricsForBehavioralDesignArXiv2020} (Section I.A), we describe in detail the theoretical limitations of earlier stable frameworks for RMPs \cite{cheng2018rmpflow,bullo2004geometric,KhatibOperationalSpaceControl1987,Peters_OpSpaceControl_AR_2008} which we know as the subclass of {\em Lagrangian} fabrics within the context of our new theory. In this paper, we show how those limitations manifest practically in direct comparisons to geometric fabrics (see Section~\ref{sec:PlanarExperiments}). In addition to their stability, geometric fabrics encode behavior as speed-invariant paths which manifests as a geometric consistency when modular parts are combined, and they retain the intuitive characteristic of acceleration-based design that makes canonical-form RMPs powerful.
\fi

\subsection{Related work}
\label{sec:RelatedWork}
\compressedversiononly
In \cite{2017_rss_system} the authors observed that even systems built on classical planning \cite{LavallePlanningAlgorithms06} or optimization \cite{RIEMORatliff2015ICRA,mukadam2017continuous,DRCIntegratedSystemTodorov2013} require a layer of real-time reactive control leveraging techniques like operational space control \cite{KhatibOperationalSpaceControl1987}. Research into Riemannian Motion Policies (RMPs) \cite{ratliff2018rmps,cheng2018rmpflow} built on these observations and proposed a behavioral design framework, embedding more globally aware behaviors into reactive control, powerful enough to develop strongly-generalizing systems\footnote{Strongly-generalizing system are systems designed and tested on a collection of validation examples that then perform robustly on an entire distribution of problems.} often circumventing standard planning architectures entirely.
\else
Much of the classical literature on robotic motion generation focuses on discrete planning queries \cite{LavallePlanningAlgorithms06}. Recent work by \cite{2017_rss_system} observed that there's a push and pull between modeling fidelity and computational expense for most modern continuous time re-planning systems, including those built on state-of-the-art optimization techniques \cite{RIEMORatliff2015ICRA,mukadam2017continuous,DRCIntegratedSystemTodorov2013}---many research implementations simplify the problem considerably to achieve the speeds they do. As such, real-time reactive motion building on classical tools such as Operational Space Control \cite{KhatibOperationalSpaceControl1987}, were proposed by \cite{2017_rss_system} as a fundamental building block of modern collaborative system design. 

A new line of research around Riemannian Motion Policies (RMPs) \cite{ratliff2018rmps,cheng2018rmpflow} addressing how to empower these reactive motion frameworks to exhibit more globally aware behaviors (such as obstacle avoidance) led to some impressive, strongly-generalizing systems\footnote{Systems designed and tested on a collection of validation examples that then perform robustly on an entire distribution of problems.} using RMPs as building blocks and circumventing standard planning architectures entirely. 
\fi

Optimization fabrics (see Section III of supplemental paper \cite{ratliffoptimizationfabricsupp}) are the culmination of that line of work into a comprehensive mathematical theory of behavioral design with rigorous stability guarantees, and geometric fabrics are their concrete incarnation. Earlier systems orchestrated RMPs in system applications using complex state machines to skirt the difficulty of designing nonlinear policies directly. These limitations motivated work on learning highly nonlinear RMPs from demonstration \cite{rana2019LearningRmpsFromDemonstration,mukadam2019RMPFusion,li2019MultiAgentRMPsArXiv}, but it proved challenging to integrate policy learning with existing RMP systems. 

A fundamental limitation of many techniques, such as operational space control \cite{KhatibOperationalSpaceControl1987,Peters_OpSpaceControl_AR_2008}, geometric control \cite{bullo2004geometric}, and geometric dynamical systems \cite{cheng2018rmpflow}, which we now understand as Lagrangian fabrics within the theory of fabrics, is that all of those systems {\em either} have fundamentally limited expressivity {\em or} they must express behavior through objective potentials (excludes velocity dependence) which creates conflicting objectives. This observation suggests that the challenges of incorporating learned RMPs stemmed from using the subclass of classical mechanical systems (a form of Lagrangian fabric). This paper studies the broader class of {\em geometric} fabric which is provably more flexible, exhibits geometrical consistency (speed-independence), and inherits rigorous stability guarantees from the theory of fabrics.

% Within the context of optimization fabrics \cite{optimizationFabricsForBehavioralDesignArXiv2020}, we now know these limitations stem from the use of classical mechanical systems (more generally the family of Lagrangian fabrics) as the behavioral model. This suggests future exploration into learning more flexible fabrics, such as geometric fabrics, are promising. \karl{should we also point out here that geometric fabrics solve potential fighting and promote speed invariance}
\section{Preliminaries}

%Geometric fabrics are a type of {\em optimization fabric} and both are derived and rigorously characterized in our companion theoretical paper \cite{OptimizationFabrics}. 
Geometric fabrics build on the theory of spectral semi-sprays (specs), which generalize the idea of modular second-order differential equations first derived and used as Riemannian Motion Policies (RMPs) in \cite{ratliff2018rmps,cheng2018rmpflow}. 
%We focus on second-order representations of policies to fully model the capabilities and dynamical considerations of physical manipulators, which themselves are modeled as second-order differential equations.
Let $\mathcal{C}$ be the $d$-dimensional configuration space of the robot.
%\footnote{By convention, we use $\q$ to denote configurations in the configuration space $\mathcal{C}$. For all other space, we denote elements of a given space using bold lower case version of the upper case script letter representing task space name, such as $\x\in\mathcal{X}$. \karl{could we just say $\q \in \mathcal{C}$ and $\x \in \mathcal{X}$ in above paragraph and omit this footer entirely}} 
Throughout this paper, we will use vector-notation describing elements of a space in coordinates. Mapped task spaces $\x = \phi(\q)$ are defined in coordinates\footnote{The spec algebra defines covariant transforms, so the behavior is independent of curvilinear changes of coordinates \cite{LeeSmoothManifolds2012}. For notational simplicity, we express our results a single choice of coordinates.} denoting $\q\in\mathcal{C}\subset\R^d$ and $\x\in\mathcal{X}\subset\R^n$ with Jacobian matrix $\J = \partial_\x\phi$, used in the relations $\xd = \J\qd$ and $\xdd = \J\qdd + \Jd\qd$.

{\em Natural-form} specs $(\M, \f)_\mathcal{X}$ represent equations of the form $\M(\x, \xd)\xdd + \f(\x, \xd) = \zero$, and their algebra derives from how these equations sum and transform under $\xdd = \J\qdd + \Jd\qd$. 
{\em Canonical-form} specs $(\M, \h)_\mathcal{X}^\mathcal{C}$ express standard acceleration-form equation $\xdd + \h(\x, \xd) = \zero$ where $\h = \M^{-1}\f$. 
For robotics applications we find it useful to additionally introduce a {\em policy-form} spec $[\M, \pi]_\mathcal{X}$ to denote the {\em solved acceleration policy} expression $\xdd = -\h(\x, \xd) = \pi(\x, \xd)$. 
%Aside from the negated $\pi = -\h$ the algebra and semantics is otherwise the same.\footnote{The notation has been fluctuating in earlier papers on RMPs \cite{ratliff2018rmps,cheng2018rmpflow}. Here we establish concretely that $(\cdot,\cdot)$ expressions represent differential equations forms, and $[\cdot,\cdot]$ denote policy semantics.} 
%The policy form makes it easier to understand the spec as an {\em acceleration policy} $\pi$ with an associated priority metric $\M$.

In practice, we usually construct a transform tree of task spaces where the specs reside. Each directed edge of the tree represents the differentiable map taking its parent (domain) to its child (co-domain). Specs populating a transform tree collectively represent a complete second-order differential equation in parts, linking a given spec to the root via the chain of differentiable maps encountered along the unique path to the root. Denoting that composed map as $\x = \phi(\q)$ as above, we can use the expressions $\xd = \J\qd$ and $\xdd = \J\qdd + \Jd\qd$ relating velocities and accelerations in the task space to velocities and accelerations in the root to derive a spec algebra that defines both how specs combine on a single space and how they transform backward across edges from child to parent
%The spec algebra enables the design of differential equations in parts using transform trees 
(see \cite{cheng2018rmpflow} for details). The tree implicitly represents a complete differential equation at the root as a sum of the parts, computed by recursive application of the spec algebra.
% the construction of transform trees of task spaces that can be populated by specs representing parts of a differential equation. 
% Through the algebraic operations of spec summation and pullback, following \cite{cheng2018rmpflow, optimizationFabricsForBehavioralDesignArXiv2020}, the tree implicitly represents a complete differential equation at the root as a sum of the parts. 

% In general, this space can be a manifold, but for simplicity of presentation, we assume it is a vector space in coordinates. Handling manifolds amounts simply to changing coordinates when necessary since the algorithms described here are known to be covariant (see \cite{OptimizationFabrics}). To best match the natural structure of physical equations and the most common context of nonlinear geometries, which are both second-order in natural, we model policies as nonlinear second-order differential equations $\qdd = \pi(\q, \qd)$ best viewed as mappings from position and velocity to acceleration (the full phase state consists of positions and velocities, and the differential equation defines how we want that state to change (acceleration)). 

%\fullversiononly
As in \cite{cheng2018rmpflow} second-order differential equations can easily be executed on fully actuated robotic systems using standard control techniques such as feedback linearization. For instance, a straightforward method is to use policies as trajectory generators (integrate forward integral curves) and follow those in the physical system using PID control as in \cite{Righetti_AR_2013}.

\section{Theoretical Development}
Here we review some of the necessary background for understanding geometric fabrics. We discuss the speed independence of {\em nonlinear geometries of paths} and how we can exploit that by {\em energizing} such geometries to conserve a type of energy called a Finsler energy which, itself, has geometric underpinnings. We then show how such energized geometries can be used to give global stability characterizations through a Hamiltonian much like in classical mechanics (the Hamiltonian can be used as a Lyapunov function). 

Importantly, we note that these energized geometries can be viewed in two complimentary ways which gives intuition around how to use them in behavioral design. First, we can view it as endowing the geometry with a priority metric derived from the energy. Or second, and equivalently, since each Finsler energy has its own associated geometry, we can view it as {\em bending} that geometry into the shape of the desired geometry using a {\em zero work modification} which preserves its metric structure. This analysis shows that these energized geometries (what we will see are {\em geometric fabrics} in the next section) can be used to optimize potential functions; later on, we show that potential functions can capture task objectives and therefore, optimizing potential functions are akin to solving task objectives while using the geometry to shape the behavior along the way.

%The following mathematical development of geometric fabrics includes discussion on nonlinear geometry, energization and energy conservation, and ultimately, proving that geometric fabrics can optimize potential functions. Later on, we show that potential functions can capture task objectives and therefore, optimizing potential functions are akin to solving task objectives.

%The theory of generalized nonlinear geometry and Finsler energy is important for the derivation of geometric fabrics given in \cite{optimizationFabricsForBehavioralDesignArXiv2020}. We give just a brief overview of key facts here sufficient for applications of geometric fabrics (see \cite{finslerGeometryForRoboticsArXiv2020} for more details.)
 
\subsection{Generalized nonlinear geometries}
\label{sec:NonlinearGeometries}
A generalized nonlinear geometry is an acceleration policy $\xdd = \pi(\x, \xd)$ for which $\pi$ has a special homogeneity property. We require that it be {\em positively homogeneous of degree 2} (HD2), which means that for any $\lambda \geq 0$ we have $\pi(\x, \lambda \xd) = \lambda^2\pi(\x, \xd)$. One can show that the HD2 property ensures the differential equation is more than just a collection of trajectories (its integral curves); it additionally has a {\em path consistency} property whereby every integral curve starting from a given position $\x_0$ with velocity $\xd_0 = \eta \widehat{\mathbf{n}}$ pointing in a given direction $\widehat{\mathbf{n}}$ (here $\eta > 0$) will follow {\em the same} path (see discourse in supplemental \cite{ratliffgeneralizednonlinearaccepted}). In particular, any variant of the differential equation of the form $\xdd = \pi(\x, \xd) + \alpha(t, \x, \xd)\xd$, where $\alpha \in\R$, will have integral curves that trace out the same paths as $\pi$. That geometric consistency property turns $\pi$ into a {\em geometry of paths}.

\subsection{Finsler energies} \label{sec:FinslerEnergies}

A Finsler energy $\Lag_e(\x, \xd)$ is a generalization of kinetic energy from classical mechanics (the classical kinetic energy $\mathcal{K} = \frac{1}{2}\xd^\tr\G(\x)\xd$ is a form of Finsler energy). Analogous to the classical case, the Euler-Lagrange equation applied to a Finsler energy defines an equation of motion $\M_e(\x, \xd)\xdd + \f_e(\x, \xd) = \zero$ where $\M_e = \partial^2_{\xd\xd}\Lag_e$ is the {\em energy (or metric) tensor} and $\f_e = \partial_{\xd\x}\Lag_e\xd - \partial_{\x}\Lag_e$ captures curvature terms (Coriolis and centripetal forces in classical mechanics). This equation matches the classical mechanical equations of motion when $\Lag_e = \mathcal{K}$, for which $\M_e(\x, \xd) = \G(\x)$.

In geometric fabrics, the energy tensor defines the policy's {\em priority metric} and the curvature terms $\f_e$ are used for stability. 
%See \cite{finslerGeometryForRoboticsArXiv2020} for an in-depth review of Finsler geometry. 
Finsler energies are Lagrangians, $\Lag_e(\x, \xd)$, that satisfy
\begin{enumerate}
    \item Positivity: $\mathcal{L}_e(\x, \xd) \geq \zero$ with equality only for $\xd = \zero$.
    \item Homogeneity: $\mathcal{L}_e(\x, \xd)$ is positively homogenous of degree 2 in $\xd$; i.e. for $\lambda \geq 0$ we have $\mathcal{L}_e(\x, \lambda\xd) = \lambda^2 \mathcal{L}_e(\x, \xd)$.
    \item Energy tensor invertibility: $\M_e = \partial^2_{\xd\xd}\mathcal{L}_e$ is invertible.
\end{enumerate}
The metric tensor $\M_e(\x, \xd)$ is in general a function of {\em velocity} as well as position, although the above homogeneity requirement enforces that $\M_e$ depends only on the {\em directionality} of the velocity ($\M_e(\x, \xd) = \M_e(\x, \widehat{\xd})$ for $\xd \neq\zero$) and not the magnitude. This enables directionally dependent priority matrices in Section~\ref{sec:design_tools} from Finsler energies.

\subsection{Energization} \label{sec:energization}
As previously discussed, any variant of the differential equation of the form $\xdd = \pi(\x, \xd) + \alpha(t, \x, \xd)\xd$ will have integral curves that trace out the same paths as $\pi$ (where $\pi$ is HD2). Given a particular $\mathcal{L}_e(\x, \xd)$ with associated $\M_e$ and $\f_e$ terms, $\xdd = \pi(\x, \xd) + \alpha \xd$ will conserve $\mathcal{L}_e$ if $\alpha$ is designed as (see Section IIIC of supplemental paper \cite{ratliffoptimizationfabricsupp} for more details)
\begin{align}\label{eqn:EnergizationTransformAlpha}
    \alpha = -(\xd^\tr\M_e\xd)^{-1}\xd^\tr\big[-\M_e\pi - \f_e\big].
\end{align}
Moreover, $\xdd = \pi(\x, \xd) + \alpha \xd$ can then be rewritten as 

\begin{align} \label{eqn:ZeroWorkEnergizationForm}
  \M_e\xdd + \f_e + \mP_e\big[-\M_e\pi - \f_e\big] = \zero,
\end{align}
where $\mP_e = \M_e\mR_{\p_e}$ and $\mR_{\p_e} = \M_e^{-1} - \frac{\xd\,\xd^\tr}{\xd^
\tr\M_e\xd}$. As discussed in Section IIIC of \cite{ratliffoptimizationfabricsupp}, $\f_f = \mP_e\big[-\M_e\pi - \f_e\big]$ is a zero-work modification term, which means that $\xd^\tr\f_f = \zero$. We rewrite (\ref{eqn:ZeroWorkEnergizationForm}) as $\M_e\xdd + \f_e + \f_f = \zero.$ As previously stated, this system will conserve $\mathcal{L}_e$, and therefore, its associated Hamiltonian, $\Ham_e$, will be conserved as well (i.e., $\dot{\Ham}_e = 0$).

\subsection{Optimization} \label{sec:optimization}
We can now optimize (minimize) a potential function $\psi(\x)$ given the previous energy-conserving system with the forced and damped variant, 
\begin{align} \label{eqn:ForcedDampedSystem}
\M_e\xdd + \f_e + \f_f = -\partial_\x\psi - \B\xd,
\end{align}
where $\B(\x, \xd)$ is positive definite. The total energy for this system (acting as our Lyapunov function) is $\Ham_e^{\psi} = \Ham_e + \psi(\x)$. The time rate-of-change of the total energy is
\begin{align}
    \dot{\Ham}_e^\psi 
    &= \dot{\Ham}_e + \dot{\psi}
    = \xd^\tr\big(\M_e\xdd + \f_e\big) + \partial_\x\psi^\tr \xd \\
    \label{eqn:TotalEnergyTimeDerivative}
    &= \xd^\tr\big(\M_e\xdd + \f_e + \partial_\x\psi\big).
\end{align}
Rewriting our forced and damped system as $\xdd = -\M_e^{-1}(\f_e + \f_f + \partial_\x\psi + \B\xd)$ and substituting yields,
\begin{align} \label{eqn:SystemEnergyDecrease}
    \nonumber
    \dot{\Ham}_e^\psi
    &= \xd^\tr\Big(\M_e\big(-\M_e^{-1}(\f_e + \f_f + \partial_\x\psi + \B\xd)\big) \\\nonumber
    &\ \ \ \ \ \ \ \ \ \ \ \ + \f_e + \partial_\x\psi\Big)\\
    \nonumber
    &= \xd^\tr\Big(-\f_e - \partial_\x\psi - \B\xd + \f_e + \partial_\x\psi\Big) - \xd^\tr\f_f \\
    &= -\xd^\tr\B\xd,
\end{align}
where all terms cancel except for the damping term. When $\B$ is strictly positive definite, the rate of change is strictly negative for $\xd\neq\zero$. Since $\Ham_e^{\psi} = \Ham_e + \psi$ is lower bounded and $\dot{\Ham}_e^{\psi}\leq\zero$, we must have $\dot{\Ham}_e^\psi = -\xd^\tr\B\xd \rightarrow \zero$ which implies $\xd\rightarrow\zero$, and therefore, $\xdd\rightarrow\zero$. Substituting $\xd = \zero$ and $\xdd = \zero$ into (\ref{eqn:ForcedDampedSystem}) yields $\partial_\x\psi = \zero$, indicating that the system has come to rest at a minimum of $\psi(\x)$.

\section{Geometric fabrics} \label{sec:GeometricFabrics}

% \nathan{
% Much of the above should be focused to satisfy the requirements of this section. Requirements: 
% 1. Finsler energy (Euler-Lagrange equation; equations of motion with $\M_e \xdd + \f_e = \zero$; $\M_e$ is positive def), 
% 2. generalized nonlinear geometry (should use notation $\xdd = \pi(\x, \xd)$; define HD2; describe as {\em acceleration policy} (mention that we refer to the policy as a {\em geometry}); forms a {\em geometry of paths}, path consistency). 
% 3. RMPs/specs (natural/policy form $(.,.)$ vs $[.,.]$, policy form uses $\pi = -\h$ (differs from canonical form by a negative sign---emphasize difference from theory paper)); context of classical mechanics and op space control. Emphasize that both policies and metrics should be velocity dependent. 
% 4. Mention that we often call {\em task} spaces just spaces, and specifically that we use the term {\em task} to refer to something we want to accomplish. 
% 5. Add algorithms for design and execution. }

Above we showed that because geometric equations define collections of path, independent of speed, we can {\em energize} them to conserve a given measure of Finsler energy (which in turn endows them with a priority metric and can be viewed as bending the corresponding Finsler geometry into the shape of the given arbitrary geometric equation). Doing so creates a theoretical context we can use to understand their global properties: when forced by a potential function (and damped), they are guaranteed to minimize the potential over the domain. The geometry influences the optimization path\footnote{The {\em optimization path} is the system trajectory generated when a fabric is forced by the negative gradient of an objective.} (the system's local behavior), but its global stability is governed by the overall convergence a local minimum of the potential function. 

Together, these properties allow us to specify tasks using potential functions (whose local minima describe task goals), and separately construct any number of behaviors by shaping the underlying geometry, while always maintaining convergence guarantees. We call such energized geometries {\em geometric fabrics}. Geometric fabrics are a form of {\em optimization fabric} (see Section III of supplemental paper \cite{ratliffoptimizationfabricsupp}) which is the broader term for this type of differential equation designed to induce {\em behavior} by influencing the optimization path of a differential optimizer. %When the fabric is {\em forced} away from its path by the negative gradient of an objective, the system is guaranteed to converge to a local optimum of the optimization problem. 
Here we describe pragmatically how to effectively design the fabric to encode a desired behavior.

A {\em forced geometric fabric} is a collection of {\em fabric terms} defined as pairs $(\Lag_e, \pi)_\mathcal{X}$ of a Finsler energy $\Lag_e(\x, \xd)$ and an acceleration policy $\xdd = \pi(\x, \xd)$. Geometric terms define the fabric while forcing terms define the objective. A {\em geometric term} is a term $(\mathcal{L}_e, \pi_2)_\mathcal{X}$ for which $\pi_2$ is an HD2 geometry. A {\em forcing term} is a term $\big(\mathcal{L}_e, -\M_e^{-1}\partial_\x\psi\big)_\mathcal{X}$ which derives its policy from a potential function. As shown in Section X, we can optimize (minimize) this potential function when damping the system. Fabric terms can be added to spaces of a transform tree \cite{cheng2018rmpflow} for the modular design of composite behaviors.

%Through the Euler-Lagrange equation $\M_e \xdd + \f_e = \zero$ (see \cite{finslerGeometryForRoboticsArXiv2020}), we 
Each fabric term defines a triple $(\M_e, \f_e, \pi)_\mathcal{X}$, where $\M_e \xdd + \f_e = \zero$ derives from the Euler-Lagrange equation applied to $\Lag_e$,
which can be viewed as two specs, a policy spec $[\M_e, \pi]_\mathcal{X}$ and a natural energy spec $(\M_e, \f_e)_\mathcal{X}$. Geometric fabric summation and pullback is, accordingly, defined in terms of the algebra of these two constituent specs. 

%In Section~\ref{sec:FinslerEnergies} we define pullback and summation operations for Finsler energies directly as well. Therefore, we 
%We can always summarize the resulting summed and/or pulled-back fabric term as its corresponding pairwise form $(\mathcal{L}_e, \pi)$. Doing so enables deriving direct expressions for the geometric fabric algebra on the pair representation. However, that's used just conceptually below, so we focus on the two-spec representation here.

Geometric fabrics are {\em unbiased} and thereby never prevent a system from reaching a local minimum of the objective. The objective, therefore, encodes concrete task goals independent of the fabric's behavior. Additionally, geometric policies are geometrically consistent speed-invariant geometry of paths, which both simplifies the intuition on how they sum and enables behavior-invariant execution speed control.

% A geometric fabric is a type of optimization fabric \cite{optimizationFabricsForBehavioralDesignArXiv2020} and acts as an unbiased optimization medium. The theory dictates we can optimize any objective over the fabric by forcing it with the negative gradient of a smooth potential and choosing appropriate amount of damping (see Section~\ref{sec:SpeedControl}).
% This theory separates the design of a task-independent nominal behavior expressed by the fabric (represented by the geometric terms) from the task-specific objective (represented by the forcing terms) 
% \karl{although, we do design geometric terms for particular tasks as well, e.g., the extractor and column attractor for the cubbies. the geometric terms can be task invariant or task dependent and we want to make as many things geometric as possible because then we get the speed invariance}.
% See Section~\ref{sec:design_tools} for an overview of specific geometric fabric designs including boundary conforming terms for modeling obstacles and joint limits. \karl{but the specifics are no long here, right?} 

The design of a geometric fabric follows the intuition of designing RMPs \cite{ratliff2018rmps}. Policy specs $[\M_e, \pi_2]_\mathcal{X}$ model both a desired behavior $\pi_2$ and a priority matrix on that behavior $\M_e$ defining how it combines with other policies as a metric-weighted average of parts. The spectrum of $\M_e$ can assign different weights to different directions in the space, and both $\pi_2(\x, \xd)$ and $\M_e(\x, \xd)$ have the flexibily of depending on both position $\x$ and velocity $\xd$.
%, and we do so in parts understanding that the final behavior will be a metric-weighted average of the parts. 
%Specifically, the acceleration policy $\pi$ should define how we want the system to accelerate in that space, and the metric $\M_e$ should be seen as an independently designed spectral weight matrix that defines how the policy should combine with other policies (colloquially, what directions in the space the policy cares about). 
Since $\M_e$ is HD0 (see Section~\ref{sec:FinslerEnergies}), geometric terms remain geometric under summation and pullback (e.g. the policy resulting from a metric-weighted average of geometries is itself a geometry).
%the metric-weighted averages of policies in sums of geometric terms
%, so for the policy specs of geometric terms, since $\pi_2$ is an HD2 geometry, the metric-weighted average implementing policy spec summation results in an acceleration policy which is, itself, HD2 and hence a generalized nonlinear geometry thereby forming a consistent geometry of paths.
The energy spec $(\M_e, \f_e)_\mathcal{X}$ of each fabric term is used only to guarantee stability during execution (see Section~\ref{sec:SpeedControl}).
Practitioners can, therefore, simply focus on designing the behavior policy specs $[\M_e, \pi_2]_\mathcal{X}$.

\section{Exploiting geometries for speed regulation}
\label{sec:speed_control}

As we did in Section~\ref{sec:energization} for energization, we will again exploit the speed independence of these geometric paths, this time to accelerate those energized systems along the direction of motion to maintain a separate measure of desired speed as faithfully as possible without violating the stability constraints outlined in Section~\ref{sec:optimization} (positive damping on the energized system). Fundamentally, this involves using the potential function to speed up by injecting energy into the system and using the damping term to slow down when desired by bleeding energy off. At times we can also explicitly inject additional energy into the system (what we call {\em boosting} the energy) as long as those energy injections are transient. For instance, we can use energy boosting to speed the system from rest quickly. By acting only along the direction of motion, the damper will not change the geometric path.

Details of our speed control methodology are given in the supplemental paper \cite{ratliffoptimizationfabricsupp}, but we review the framework here. Let $\pi_0(\x,\xd)$ be a geometry, $\M_e$ be its metric induced through energization via Finsler energy $\Lag_e$, and $\psi(\x)$ be a forcing potential. The speed of the forced system can be regulated using an acceleration along the direction of motion $\alpha_\mathrm{reg}\xd$ using
\begin{align} \label{eqn:FundamentalsOfSpeedControl}
    \xdd = -\M_e^{-1}\partial_\x\psi(\x) + \pi_0(\x, \xd) + \alpha_\mathrm{reg} \xd.
\end{align}
By Theorem~III.18 of \cite{ratliffoptimizationfabricsupp}, as long as $\alpha_\mathrm{reg} < \alpha_{\Lag_e}$ where $\alpha_{\Lag_e}$ is the energization coefficient (see Equation~\ref{eqn:EnergizationTransformAlpha}), the system is equivalent to the original energized system with strictly positive damping coefficient. This result can be derived simply by using a velocity aligned damper $-\beta \xd$ on the energized system and rearranging the terms by collecting the energization term and velocity damper together (both of which are accelerations along the direction of motion). The above constraint then derives from the original conditions required for stability.

\section{Execution and algorithms}
\label{sec:SpeedControl}

Once the forced geometric fabric is designed, one can transform it by accelerating and decelerating along the direction of motion using the methodology outlined in Section~\ref{sec:speed_control} (see also Section~\ref{subsec:damping_details})
to maintain a given measured of {\em execution energy} (e.g. speed of the end-effector or joint speed through the configuration space).
The geometric consistency of the fabric ensures the behavior remains consistent despite these speed modulations. Many numerical integrators are appropriate for integrating the final differential equation. We find Euler (1ms time step) or fourth order Runge-Kutta (10ms time step) exhibit a good trade-off between speed and accuracy.

% We have found that a good balance between integration speed and accuracy can be met with explicit Euler with a 1 ms time step or fourth order Runge-Kutta with a maximum time step of 10 ms. Other integration methods are useful as well like symplectic Euler or trapezoidal rule.

% The final step in executing a forced geometric fabric is speed modulation. We can drive the system through the fabric at any desired speed, as long as a specific lower bound on the required damping is met. Since geometric paths are independent of speed, speed control is particularly useful for geometric fabrics. We follow the speed control methodology outlined in \cite{optimizationFabricsForBehavioralDesignArXiv2020} with specific functional forms described in Section~\ref{subsec:damping_details}.

% The obtained differential equation can be integrated with respect to time using various integration methods. We have found that a good balance between integration speed and accuracy can be met with explicit Euler with a 1 ms time step or fourth order Runge-Kutta with a maximum time step of 10 ms. Other integration methods are useful as well like symplectic Euler or trapezoidal rule. 

In full, we design a system in three parts (using a transform tree \cite{cheng2018rmpflow} of task spaces): (1) design the underlying behavioral fabric, (2) add a driving potential to define task goals, (3) design an execution energy for speed control.

%\noindent\fbox{
%\parbox{\columnwidth}{

\fullversiononly
\vspace{2mm}
\noindent Design of a forced geometric fabric:
\begin{enumerate}
    \item Construct a transform tree.
    \item Populate its nodes with fabric terms. Most terms are geometric terms and just a few (often only one) are forcing terms.
    \item Add execution energy specs to the tree as needed to describe the execution energy we want to control.
\end{enumerate}
Execution of the forced fabric with speed control:
\begin{enumerate}
    \item Forward pass: Populate the nodes with current state from the root to the leaves.
    \item Backward pass: Evaluate the specs an pull them to root in separate channels, an {\em energy channel} for the geometric terms' energy specs, a {\em policy channel} for the geometric terms' policy specs, an {\em execution energy channel} for the execution energy specs, and a {\em forcing policy channel} for the forcing terms' policy specs. Add the forcing term's energy specs to the energy channel.
    \item Use four channels' combined root results to calculate the final desired acceleration using speed control.
\end{enumerate}
\else % Short version
\vspace{2mm}
\noindent {\bf Design of a forced geometric fabric:}\\ 
(1) Construct a transform tree.\\
(2) Populate its nodes with fabric terms. \\
% Most terms are geometric terms and just a few are forcing terms (often only one).
(3) Add execution energy specs to the tree as needed to describe the execution energy we want to control.

\vspace{2mm}
\noindent {\bf Execution of the forced fabric with speed control:}\\
(1) Forward pass: Populate the nodes with the current state from the root to the leaves.\\
(2) Backward pass: Evaluate the specs and pull them to root in separate channels, an {\em energy channel} for the geometric terms' energy specs, a {\em policy channel} for the geometric terms' policy specs, an {\em execution energy channel} for the execution energy specs, and a {\em forcing policy channel} for the forcing terms' policy specs. Add the forcing term's energy specs to the energy channel.\\
(3) Use the four channels' root results to calculate the final desired acceleration using speed control.

 Fig. \ref{fig:tree} shows a transform tree of different spaces with four differently colored channels pass energies, policies, forcing policies, and execution energies backwards through the tree. 

\setcounter{figure}{1}
\begin{figure}[!t]
  \centering
  \includegraphics[width=0.8\linewidth]{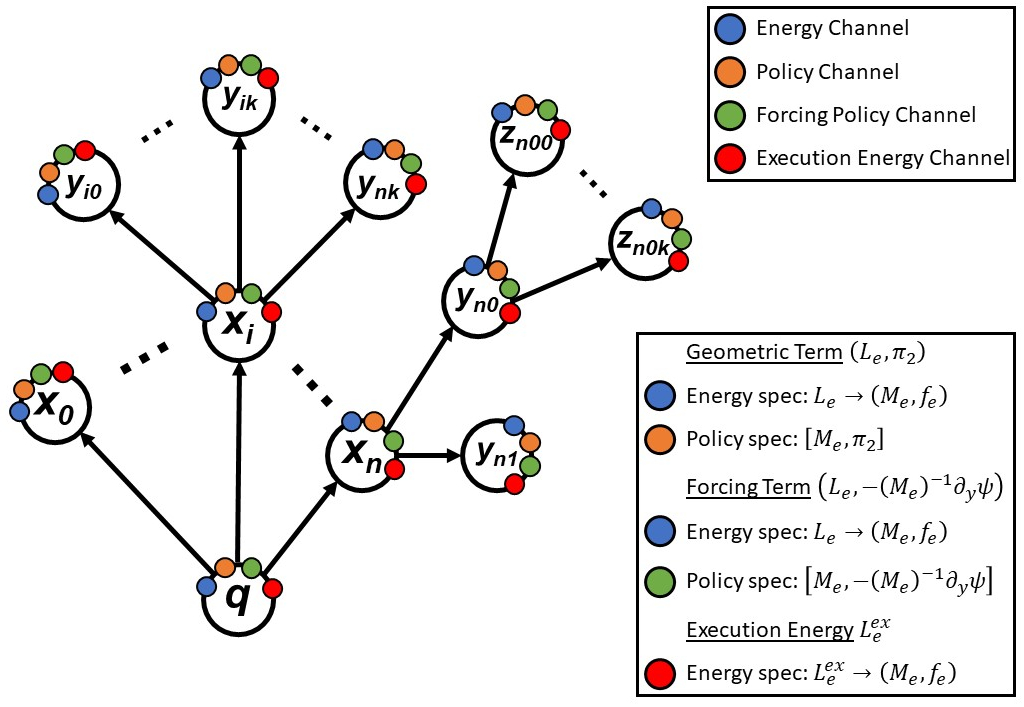}
  \vspace{-4mm}
  \caption{Forced geometric fabric based on a transform tree of task spaces with four channels pass energies, policies, forcing policies, and execution energies.}
  \vspace{-4mm}
  \label{fig:tree}
\end{figure}

\fi

\section{Concrete Design Tools}
\label{sec:design_tools}
%\subsection{Prioritization of Geometries}
Geometric fabrics follow acceleration-based design principles captured in the original canonical-form RMPs \cite{ratliff2018rmps}. As described in Section~\ref{sec:GeometricFabrics}, a geometric fabric is a pair $(\Lag_e,\pi_2)_\mathcal{X}$ characterizing two specs, an energy spec and a geometry spec. The energy spec captures stability information, while the geometry spec captures behavior. Behavioral design focuses on constructing the latter, using the class of HD2 geometries to model $\pi_2$ and deriving $\M_e$ as the energy tensor from $\Lag_e$.

When geometric fabrics are summed $\sum_i (\Lag_e^{(i)}, \pi_2^{(i)})_\mathcal{X}$, the combined fabric's geometry spec (capturing its behavior) $\sum_i (\M_e^{(i)}, \pi_2^{(i)})_\mathcal{X} = \big(\wt{\M}_e, \wt{\pi}_2\big)$ is a metric-weighted average of the contributing geometries $\wt{\pi}_2 = \big(\sum_i \M_e^{(i)}\big)^{-1}\sum_i\M_e^{(i)}\pi_2^{(i)}$, prioritized by the total metric $\wt{\M}_e = \sum_i \M_e^{(i)}$. When populating a transform tree, this intuitive combination rule is applied recursively at each node. Designers need only focus on intuitively creating modular acceleration policies (as HD2 geometries) in the different spaces and prioritizing them with metric tensors (from Finsler energies).

\subsection{Construction of HD2 Geometries}
\label{subsec:geometry}
An HD2 geometry is a differential equation $\xdd + \h_2(\x, \xd) = \zero$ where $\h_2$ is HD2 (see Section~\ref{sec:NonlinearGeometries}), which we usually denote in {\em policy} form $\xdd = -\h_2(\x, \xd) = \pi_2(\x, \xd)$. Constructing an HD2 geometry is straightforward given the following rules of homogeneous functions: (1) a sum of HD2 functions is HD2; (2) multiplying homogeneous functions adds their degrees (denoting an HD$k$ function as $f_k$, examples are $f_2f_0 = f_2$, $f_1f_1 = f_2$, etc.). For instance, a simple way to design an HD2 geometry is to choose an HD0 policy $\pi_0(\x)$ that depends only on position and form $\pi_2(\x, \xd) = \|\xd\|^2\pi_0(\x)$ by scaling it by $\|\xd\|^2$. $\pi_0(\x)$ can be chosen as the negative gradient of a potential $\pi_0(\x) = -\partial_\x\psi(\x)$. %(see, for instance, Section~\ref{sec:Vortices}).

%Let

% \nathan{Rules for homogeneous combination: all terms need to be HD2, products of terms multiply their HD2, so HD1*HD1 = HD2; HD0*HD2=HD2. Then describe potential as a special type of HD2*HD0 implementation; vortex is another version (ref below).}

% In general, geometries are functions of position and velocity and the only requirement is that the function is homogeneous of degree 2 in velocity. However, strategies for easily constructing geometries exist. For instance, a formulaic approach to constructing a geometry is to design a potential function in task space $\x$ as $\psi(\x)$. Taking the gradient of this potential with respect to position and pre-multiplying by the inner product of $\xd$ will create a geometry of the form,

% \begin{align}
%     \h_{2}(\x, \xd) = \|\xd\|^2 \partial_\x \psi(\x).
% \end{align}

% This geometry will effectively create a force in the negative direction of the potential gradient, scaled by velocity. The homogeneity condition is easily met with the inclusion of $\|\xd\|^2$, i.e., $\|\lambda \xd\|^2 = (\lambda\xd)^\tr(\lambda\xd) = \lambda^2  \xd^\tr\xd = \lambda^2\|\xd\|^2$. Overall, geometric behavior with this design generates forces that push towards the minima of $\psi(\x)$. However, this design is not the only way to build geometries, and others can be constructed without any concept of pushing towards potential minima.

\subsection{Acceleration-based Potentials}
\label{subsec:acceleration_potential}

It is often most intuitive to design a geometric fabrics' forcing potential as a forcing {\em spec} $\mathcal{F} = [\M_f, \pi_f]_\mathcal{X}$ in {\em policy form} so it's treated intuitively as another acceleration policy averaged into the final metric weighted average. This policy $\pi_f$ must implicitly express a {\em forcing potential} $\psi_f(\x)$ whose negative gradient is given by $-\partial_\x\psi_f(\x) = \M_f\pi_f$. When designing $\mathcal{F}$, $\M_f$ and $\pi_f$ must remain theoretically compatible in that sense. Following Appendix D.4 in \cite{cheng2018rmpflowarxiv}, we advocate choosing $\pi_f = -\nabla_\x \psi_\mathrm{acc}(\x)$ where $\psi_\mathrm{acc}$ is a potential function that is spherically symmetric around its global minimum expressing the acceleration policy directly as its negative gradient. Any metric $\M_f(\x)$ is theoretically compatible if it is also spherically symmetric around the same global minimum point. Note that position-only metrics are Riemannian and derive from Finsler energies of the form $\Lag_e = \frac{1}{2}\xd^\tr\M_f(\x)\xd$.

\subsection{Speed control}
\label{subsec:damping_details}

We follow the speed control methodology outlined in Section~\ref{sec:speed_control}. Specifically, that entails using $\alpha_\mathrm{reg} = \alpha_\mathrm{ex}^\eta - \beta_\mathrm{reg}(\x, \xd) + \alpha_\mathrm{boost}$ in $\xdd = -\M_e^{-1}\partial_\x\psi(x) + \pi_0(\x, \xd) + \alpha_\mathrm{reg} \xd$ with $\beta_\mathrm{reg} = s_\beta(\x) B + \underline{B} + \max\{0, \alpha_\mathrm{ex}^\eta - \alpha_{\Lag_e}\}$ 
% (see \cite{optimizationFabricsForBehavioralDesignArXiv2020} for details). 
$\underline{B} > 0$ is a (small) baseline damping, the $B > \underline{B}$ is a larger damping coefficient for succinct convergence. The switch $s_\beta(\x)$ turns on close to the target:
\begin{align}
\label{eqn:damping_gate}
 s_\beta(\x) = \frac{1}{2} \Big(\tanh\big(-\alpha_\beta (\|\x\| - r) \big) + 1\Big) 
\end{align}
where $\alpha_\beta \in \mathbb{R}^+$ is a gain defining the switching rate, and $r \in \mathbb{R}^+$ is the radius where the switch is half-way engaged. Denoting the desired execution energy as $\Lag_e^\mathrm{ex,d}$, we use the following policy for $\eta$
\begin{align}
\label{eqn:energy_gate}
    \eta = \frac{1}{2} \Big(\tanh\big(-\alpha_\eta (\Lag_e^\mathrm{ex} - \Lag_e^\mathrm{ex,d}) - \alpha_\mathrm{shift} \big) + 1\Big) 
\end{align}
where $\alpha_\eta,\alpha_\mathrm{shift} \in \mathbb{R}^+$ adjust the rate and offset, respectively, of the switch as an affine function of the speed (execution energy) error. Finally, $\alpha_\mathrm{boost}$ is modeled as $\alpha_{boost} = k \: \eta \big(1-s_\beta(\x)\big)\frac{1}{\|\xd\| + \epsilon}$, where $k \in \mathbb{R}^+$ is a gain that directly sets the desired level of acceleration, $\eta$ (from above) sets $\alpha_{boost}=0$ when the desired speed is achieved, and $1-s_\beta(\x)$ sets $\alpha_{boost}=0$ when the system is within the region of higher damping. The normalization by $\|\xd\|+\epsilon$ ensures that $\alpha_{boost}$ is directly applied along $\hat{\xd}$ with a very small positive value for $\epsilon$ to ensure numerical stability. This overall design injects more energy into the system when $-\alpha_{boost} < \alpha_\mathrm{ex}^\eta - \alpha_{\Lag_e}$. Since this injection occurs for finite time, the total system energy is still bounded. The additional switches ensure that the system is still subject to positive damping, guaranteeing convergence.

\section{Planar Manipulator Experiments} \label{sec:PlanarExperiments}
% \nathan{Make note somewhere that we build these systems up layerwise following the empirically validated approach outlined in the arxiv paper. (We're removing the particle experiments due to lack of space.)} 

\begin{figure}[!t]
  \centering
  \includegraphics[width=0.8\linewidth]{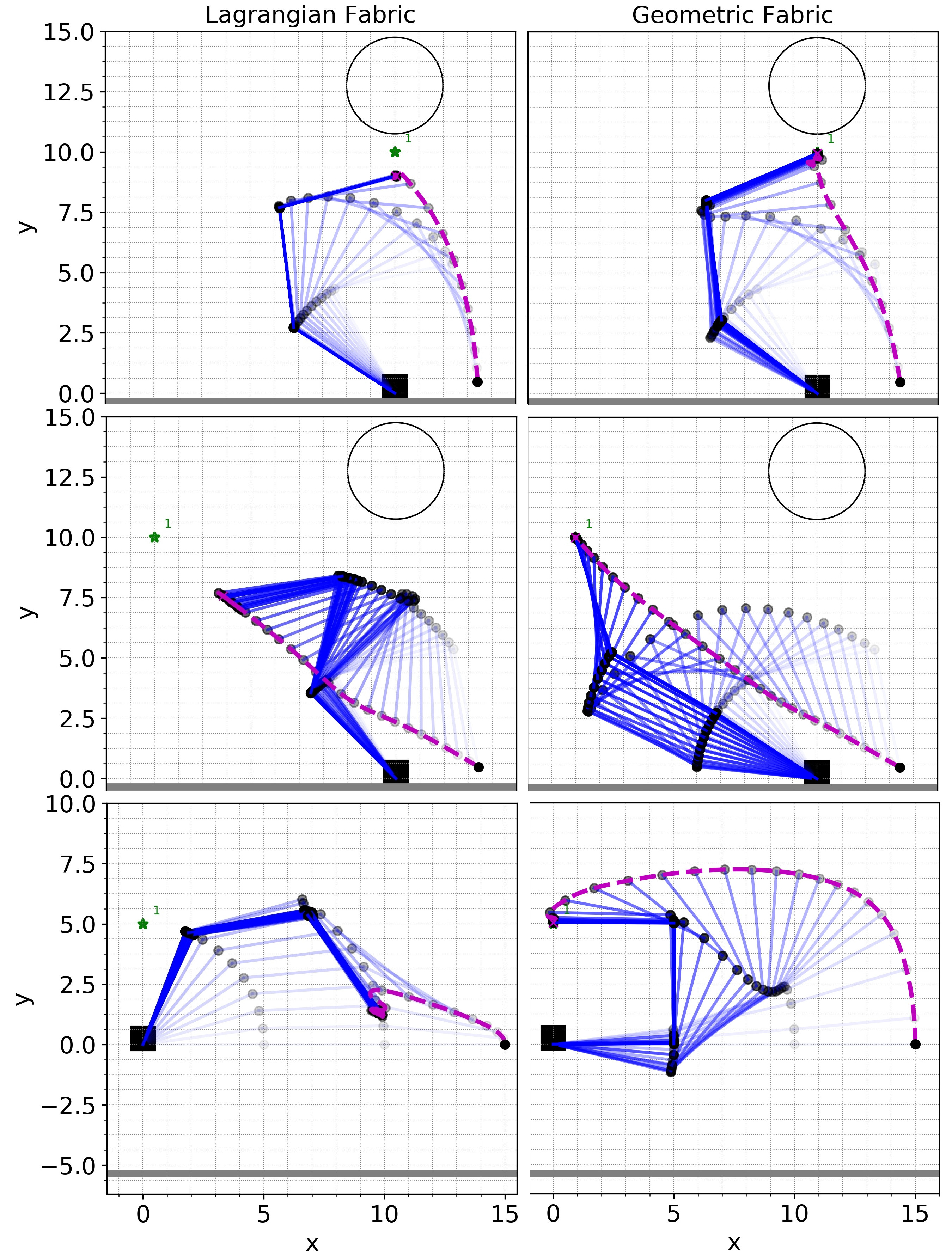}
  \vspace{-4mm}
  \caption{Planar arm experiment results for Lagrangian and geometric fabrics, where targets are marked as green stars, and the arm's alpha transparency ranges from light at the beginning of the trajectory to dark at the end.}
  \label{fig:gf_vs_lds_planar}
  \vspace{-4mm}
\end{figure}

In this section, we present a set of controlled experiments on a 3-dof planar arm to demonstrate the limitations of Lagrangian fabrics (as discussed in Section~\ref{sec:RelatedWork}) and show how geometric fabrics overcome them. 

\textbf{Lagrangian v.s. Geometric:} Denoting Lagrangian fabric terms as $(\Lag_e, \pi_l)_\mathcal{X}$ we design Lagrangian fabric acceleration policies as $\pi_l = -\partial_\x\psi$ using acceleration-based potentials $\psi$. The corresponding geometric fabric terms are $(\Lag_e, \pi_g)$ where $\pi_g = \|\xd\|^2\pi_l$, converting the position-only acceleration policies of the Lagrangian fabric into HD2 geometries for a {\em geometric} fabric simply by multiply by $\|\xd\|^2$. Lagrangian terms are actually all {\em forcing terms} per the categorization in Section~\ref{sec:GeometricFabrics}. In contrast, geometric fabric terms add to the underlying fabric and are therefore inherently unbiased.
Our experiments consistently show that the Lagrangian forcing terms can easily be tuned to compete with one another, preventing convergence to the task goal. In contrast, when converted to geometries, these same terms work collaboratively with one another to {\em influence} the optimization behavior without affecting task convergence.

All task goals are point targets for the end-effector expressed as simple acceleration-based attractor potentials (see Section~\ref{subsec:acceleration_potential}). 
\noindent Our three experiments are to: 
\begin{enumerate}
\item Reach toward a target close to an obstacle (we expect the obstacle and task terms to conflict). 
\item Reach toward a target in an ``awkward'' configuration (we expect the redundancy resolution and task terms to conflict). 
\item Reach toward a target at an ``extreme'' configuration (we expect the joint limit terms and task terms to conflict).
\end{enumerate}
Figure~\ref{fig:gf_vs_lds_planar} shows that in all cases, Lagrangian fabric (forcing) terms conflict with the task (forcing) term preventing convergence to the goal. In contrast, the analogous geometric fabrics are fundamentally unbiased and enable task convergence while still exhibiting the desired behavioral influence.

Below we describe each fabric term, focusing on describing their geometric fabric form; Lagrangian fabric terms are derived simply by dropping the velocity scaling factor $\|\xd\|^2$ using the relation $\pi_g = \|\xd\|^2\pi_l$.

% \nathan{Is the information in the Point Attraction and Circular Object Repulsion consistent with the above. Perhaps remove or merge up there? I think this entire section can effectively just be the content in the paragraph above and the experiments section (C).}

% \subsection{End-effector Attraction}
% \label{subsec:point_attraction}
\textbf{End-effector Attraction:} The behavior for pulling the end-effector toward a target is constructed as follows. The task map is $\x = \phi(\q) = \q - \q_d$, where $\q$, $\q_d \in \mathbb{R}^2$ are the current and desired particle position in Euclidean space. The metric is designed as
\begin{align}
\label{eq:potential_priority}
\G_\psi(\x) = (\widebar{m} - \underline{m}) e^{-(\alpha_m \|\x\|) ^ 2} I + \underline{m} I.
\end{align}
where $\widebar{m}$, $\underline{m}$ are the upper and lower isotropic masses, respectively, and $\alpha_m$ is a constant scalar. The acceleration-based potential gradient is $\partial_\q \psi(\x) = M_\psi(\x) \partial_\q \psi_1(\x)$, where
\begin{align}
\label{eq:attractor_potential}
\psi_1(\x) = k \left( \|\x\| + \frac{1}{\alpha_\psi}\log(1 + e^{-2\alpha_\psi \|\x\|} \right)
\end{align}
where $k$ and and $\alpha_\psi$ are constant scalars. Since our task is to ultimately reach our target location, this behavior is a forcing geometric fabric term with $\pi = -\partial_\q \psi_1(\x)$.

% For Lagrangian Fabrics, the policy is created from the Lagrangian, $\Lag = \xd^T \G_\psi(\x) \xd - 
% \psi(\x)$. For geometric fabrics, the policy is constructed as $\xdd = -\partial_\x \psi_1(\x)$ and weighted by $\M_\psi(\x)$ from the Finsler energy, $\Lag = \xd^T \G_\psi(\x) \xd$.

% \subsection{Circular Object Repulsion}
% \label{subsec:circular_object_repulsion}
\textbf{Circular Object Repulsion:} Collision avoidance with respect to a circular object is constructed as follows. The task map is $x = \phi(\q) = \frac{\|\q - \q_o\|}{r} - 1$, where $\q_o$ is the origin of the circle and $r$ is its radius. The metric is defined as $G_b (x, \dot{x}) = s(\dot{x}) \frac{k_b}{x^2}$, where $k_b$ is a barrier gain and $s(\dot{x})$ is a velocity-based switching function. Specifically, $s(\dot{x}) = 1$ if $\dot{x} < 0$ and $s(\dot{x}) = 0$, otherwise. 
% For these experiments, $k_b = 20$.
The acceleration-based potential gradient is $\partial_\q \psi_b(x) = M_b (x) \partial_\q \psi_{1,b}(x)$ with $\psi_{1,b}(x) = \frac{\alpha_b}{2 x^8}$, where $\alpha_b$ is the barrier gain.
% and $\alpha_b = 1$ for these experiments. 

% \subsection{Limit Avoidance}
% \label{subsec:limit_avoidance}

% \nathan{The information here is probably redundant with the below description of the planar robot. We need to decide where to describe which geometries. Would it make sense to factor out all the descriptions into a single place which collects the detailed descriptions of the geometries used in the experiments? Then each of the experiments can just reference which they use.}

\textbf{Limit Avoidance:} To describe the joint limit avoidance policy, we use two 1D task maps, $x = \phi(q_j) = \bar{q}_j - q_j \in \R_+$ and $x = \phi(q_j) = q_j - \underline{q}_j \in \R^+$, for each joint to capture the distance between the current joint position and its lower and upper boundaries, respectively, where $\bar{q}_j$ and $\underline{q}_j$ are the upper and lower limits of the $j^{th}$ joint. 
The 1D metric $G_l$ is defined as $G_l(x, \dot{x}) = s(\dot{x})\frac{\lambda}{x}$, where $s(\dot{x})$ is a velocity-based switching function. Specifically, $s(\dot{x}) = 1$ if $\dot{x} < 0$ and $s(\dot{x}) = 0$, otherwise. Effectively, this removes the effect of the coordinate limit geometry once motion is orthogonal or away from the limit.
The acceleration-based potential gradient, expressed as $\partial_q \psi_l(x) = M_l(x) \partial_\q \psi_{1,l}(x)$, is designed with
\begin{align}
\label{eq:limit_potential}
\psi_l(x) = \frac{\alpha_1}{x^2} + \alpha_2 \log \left(e^{-\alpha_3(x - \alpha_4)} + 1 \right)
\end{align}
where $\alpha_1$, $\alpha_2$, $\alpha_3$, and $\alpha_4$ are constant scalars.
% \noindent where $\psi_l(\x) \to \infty$ as $\x \to 0$ and $\psi(\x) \to 0$ as $\x \to \infty$. Moreover, $\alpha_1$, $\alpha_2$, $\alpha_3$, $\alpha_4 \in \mathbb{R}^+$ where, $\alpha_1$ and $\alpha_2$ are gains that control the significance and mutual balance of the first and second terms. $\alpha_3$ controls the sharpness of the smooth rectified linear unit (SmoothReLU) while $\alpha_4$ offsets the SmoothReLU. 

% For Lagrangian fabrics, the policy is created from $\Lag = \G_l(\x, \dot{\x}) \dot{\x}^2 - 
% \psi_l(\x)$. For geometric fabrics, the policy is $
% \ddot{\x} = - \dot{\x}^2 \partial_{\x} \psi_{1,l}(\x)$. This policy is weighted by $\M_l(\x, \dot{\x})$ from the Finsler energy, $\Lag = \xd^T \G_l(\x, \dot{\x}) \xd$.

% \subsection{Default Configuration}
% \label{subsec:default_config}
\textbf{Default Configuration:} The task map for default configuration policy is defined as $\x = \phi(\q) = \q - \q_0$, where $\q_0$ is a nominal configuration, and it usually represents the robot ``at ready". The metric $\mathbf{G}_{dc}$ is simply an identity matrix scaled by a constant $\lambda_{dc}$, $\mathbf{G}_{dc} = \lambda_{dc} I$. The acceleration-based potential gradient is defined with Eq.~\ref{eq:attractor_potential}.

\textbf{Experiment Results:} The end-effector attraction, default configuration and joint limit avoidance are applied to all three experiments, and obstacle avoidance is applied to the first two experiments. The nominal default configuration in all three experiments is $[\frac{\pi}{2}, -\frac{\pi}{2}, -\frac{\pi}{2}]$, where positive sign defines counter clock-wise rotation, and vice versa. The upper and lower joint limits for each joint are set as $\pm{(\pi)}$ in the first two experiments and $\pm{(\frac{\pi}{2}+0.01)}$ in the third experiment. 

The experiment results are shown in Fig.~\ref{fig:gf_vs_lds_planar}, From the top row to the bottom row, it shows the results of experiments 1 to 3. As seen in the left column, no target is reached with Lagrangian fabrics due to the fighting potentials. As seen in the right column, all three targets are reached with geometric fabrics as shown in the right column.

% \fi % end full version section 
\section{Franka Arm Experiments}

% Significance is to show GF scales to more complex settings. Should be pleasing to the robot crowd. Reaching into randomly selected cubbies across three different cubby sets. Shows collision avoidance, navigation into and out of cubbies, pleasing arm configurations, etc.

\begin{figure}[!t]
  \centering
  \includegraphics[width=0.8\linewidth]{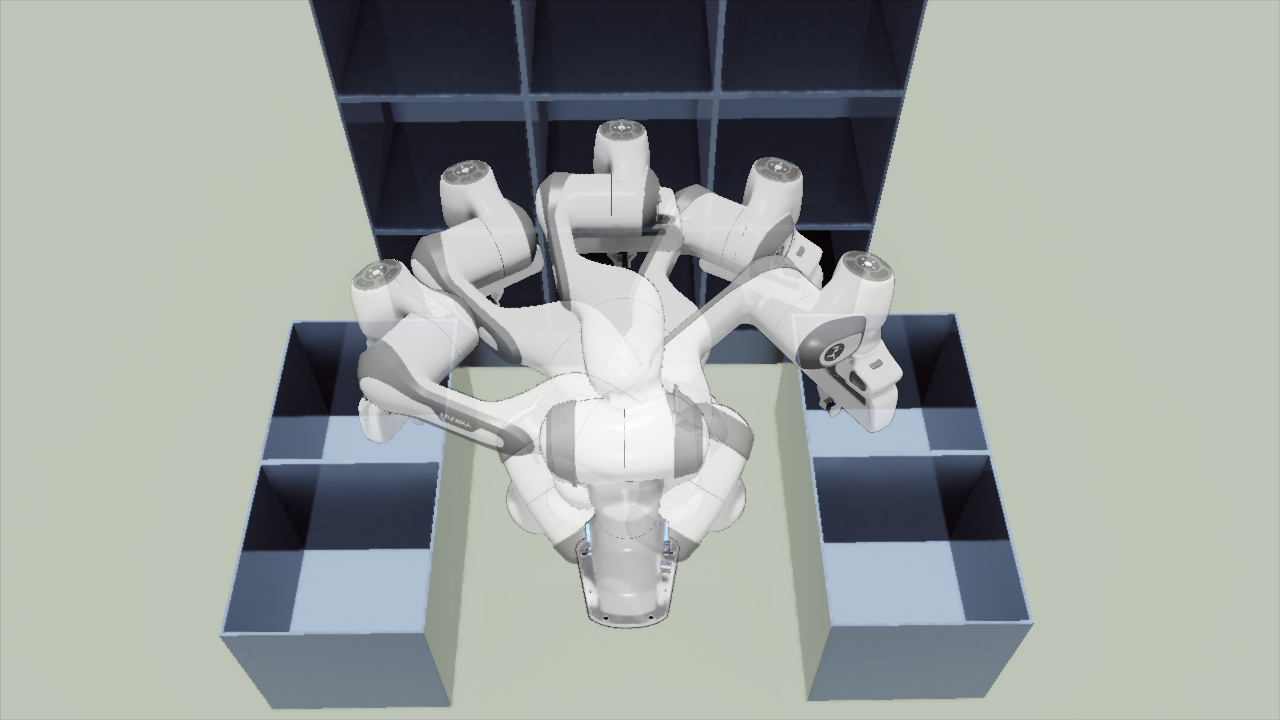}
  \vspace{-4mm}
  \caption{Five default configurations for the Franka arm.}
  \vspace{-4mm}
  \label{fig:cubby_scene}
\end{figure}

\fullversiononly

In this section, we run experiments on a 7-dof Franka arm, and show that geometric fabric scales to more complex settings. The experiments are performed in the environment shown in Fig.~\ref{fig:cubby_scene}, where there are nine cubbies in front of the Franka arm, and two cubbies on each side of the arm. The length, width and depth of each cubby are 0.3m. The task objective is to continuously move the end effector to randomly selected targets located in different cubbies. To accomplish this task, we build the geometric fabric up layerwise as follows. 

\subsection{First Layer}
\label{subsec:first_layer}
In the first layer of the geometric fabric, we design policies that generates the nominal behavior in the free space where no cubbies exists, and the objective is simply to move end effector to targets without exceeding joint limits or having redundancy issue. This can be done by adding the target attraction, the joint limit avoidance, and the default configuration, as described in \ref{subsec:target_attraction}, \ref{subsec:limit_avoidance}, and \ref{subsec:default_config}, respectively to the first layer. Instead of using one default configuration, here we have five default configuration candidates, and they are configurations with end-effector points to the left, the front left, the front, the front right and the right. The one has the end-effector pointing to the front is shown in Fig~\ref{fig:cubby_scene}. We select the default configuration from the five candidates based on the proximity of the end-effector position in each default configuration and the target. To encourage straight-line motion in the end-effector task space and also to funnel to the attractor, we add another target attraction policy to the first layer. This target attraction police is similar to the one descried in \ref{subsec:target_attraction}, but with a slight modification since it is formulated as a geometry, $
\ddot{\x} = - \dot{\x}^2 \partial_{\x} \psi_{1}(\x)$. 

\subsection{Second Layer}
\label{subsec:second_layer}
Once we get the desired nominal behavior, we design the second layer of the geometric fabric with a cubby extraction policy that can extract the end effector out of the current cubby, and keep it away from the cubbies until it is in front of the target cubby. 

The cubby extraction policy is described with two task maps. The first one is defined as the distance between the end effector and the front plane of the cubby, $\x_1=\phi_1(\y)$, where $\y$ $\in \mathbb{R}^3$ is the current end-effector position in Euclidean space. The second task map is defined as the distance between the end-effector and a cylinder, whose center line is orthogonal to the front plane of the target cubby while passing by the center of the target cubby, $\x_2=\phi_2(\y)$, , where $\y$ $\in \mathbb{R}^3$ is the current end-effector position in Euclidean space. The radius of the cylinder is half of the cubby size, which is $0.15m$. The priority metric is defined in the way such that the priority vanishes if the end-effector is either more than $d=0.1m$ away from the front plane of the cubby (outside of the cubby) or it is inside of the cylinder aligned with the target cubby, specifically: 
\begin{align}
\label{eq:M_bw}
\G_w (\y) = s(\x_1)((\widebar{m} - \underline{m})\textbf{T}(\x_2) + \underline{m}) I.
\end{align}
where $s(\x_1)$ is a range-based switching function. Specifically, $s(\x_1) = 1$ if $\x_1<d$ and $s(\x_1) = 0$, otherwise. $\textbf{T}(\x_2)=\tanh(-(\alpha_m \|\x_2\|) ^ 2 + 1)$, where $\widebar{m}$, $\underline{m} \in \mathbb{R}^+$ are the upper and lower isotropic masses, respectively, and $\alpha_m \in \mathbb{R}^+$ defines the width of the transition between $0$ and $1$.  The acceleration-based potential gradient and policy formulation are the same as described in \ref{subsec:limit_avoidance}

To help pulling the end-effector to the target inside a cubby, we design another target attraction policy in this layer, which is similar to the one defined in Section \ref{subsec:target_attraction}, but with some modification in both the priority metric and the policy formulation. The metric is designed with the same formula as in Eq.~\ref{eq:M_bw} but complimentary to the metric for wall repulsion policy, meaning the priority remains zero until it enters the cylinder aligned with the target cubby.

\subsection{Third Layer}
\label{subsec:third_layer}
Once we get the desired behavior, we add a cubby collision policy to the third layer of the fabric for collision avoidance. The cubby collision policy is the same as the circular obstacle avoidance defined in Section \ref{subsec:obstacle_avoidance}, except for a slightly different definition of the task map. To describe the cubby collision policy, we use a one-dimensional task space $\x=\phi(\y)$ \nathan{This equation reads like a multi-dimensional space. Use $\x$ as the point on the robot and $s = d(\x)$ as a representation of the task space (no bold). (I've seen $s\in\R$ commonly in physics texts to denote distance.)} to capture the minimum distance between a point on the arm and the cubby, where $\y$ $\in \mathbb{R}^3$ is the coordinates of a point on the robot. If the point is inside of a cubby, the minimum distance is between the point and one of the walls of the cubby. Otherwise, the minimum distance is between the point and one of the edges of the cubby. \nathan{Just say ``distance to cubby''. Distances functions like this are common in robotics.}

\subsection{Cubby Experiment Results}
We generate a random set of targets, and run simulation with three different settings. First, we run simulation with only the first layer of the geometric fabric. Next, we run simulation with the addition of the second layer to the fabric. Last, we run simulation with the geometric fabric with all three layers. We demonstrate the simulation results with video clips presented in our website \url{https://sites.google.com/nvidia.com/geometric-fabrics}. As we designed, the first layer of the geometric fabric generates the nominal behavior, which simply moves the end-effector to each target in the free space. With the default configuration policy, the robot always moves around its front side. \nathan{Make sure we emphasize this global navigation problem above.} In the first video clip, we can observe that the robot moves from the cubby on its left side to the cubby on its right side around its front instead of its back. With the addition of the second layer, the geometric fabric generates motion that moves the end-effector out of the current cubby first, and then approaches to the target after it enters the target cubby, as shown in the second video clip. With all three layers, the geometric fabric generates motion that moves the end-effector to targets located in different cubbies continuously while preventing the robot from colliding with the cubby, as shown in the third video clip.      
 
\else % Compressed version

These experiments represent a global end-effector navigation problem common in many logistics, collaborative, and industrial settings, wherein a robot must autonomously interact with cubbies (bins). We demonstrate the effectiveness of the layer-wise construction of a geometric fabric that enables the robot to reach into and out of three sets of cubbies (see Fig.~\ref{fig:CubbyResults}). There are six reachable cubbies in front and two on either side of the Franka arm. The length, width, and depth of each cubby are $0.3\: m$. The following discusses the fabric layers, where every new layer rests upon the previous, fixed layers, mitigating design and tuning complexity.

\subsection{Behavioral Layers}
We construct the behavior in three layers: 1) global cross-body navigation; 2) heuristic geometries for moving in and out of cubbies; 3) collision avoidance. Layers 1 and 3 are general and can be used in many contexts; Layer 2 is a common heuristic, although implemented here using task specific considerations.
Each policy is an HD2 geometry of the form $\xdd = - \|\xd\|^2 \partial_{\x} \psi(\x)$. Policies are weighted by $\M_e(\x, \xd)$ from the Finsler energy, 
$\Lag_e = \xd^T \G(\x, \xd) \xd$. 
$\psi(\x)$ and $\G(\x, \xd)$ are defined as follows.
\subsubsection{Layer 1}
The first layer creates a baseline geometric fabric designed for {\em global}, cross-body, point-to-point end-effector navigation absent obstacles. We construct end-effector attraction, joint limit avoidance, and default configuration geometries (see Section~\ref{sec:PlanarExperiments}). Five separate default configurations are created to cover different regions of the robot (see Fig~\ref{fig:cubby_scene}). This behavior controls robot posture and resolves manipulator redundancy.

\subsubsection{Layer 2}
This layer enables the end-effector to extract from and enter into any cubby (ignoring collision).

%An {\em extraction} geometry uniformly accelerates the end-effector out of {\em any} cubby {\em except} the target cubby. The extraction geometry's priority vanishes as it leaves the cubby, constraining its governance to only the cubby. In a cylindrical region emanating from a target cubby, we modulate the metric so as to turn off that extraction geometry and tie that to the simultaneous activation an additional attraction geometry.

\textbf{Cubby Extraction:} This uses two task maps that are the distance between: 1) the end-effector and the {\em front plane} of the cubby,  $y_1=\phi_1(\x)=sgn*\|\x_f-\x\|$, $sgn=-1$ if the end-effector is inside of the cubby, $sgn=1$, otherwise, where $\x, \x_f \in \mathbb{R}^3$ are the end-effector position and its orthogonal projection onto the front plane, respectively; and 2) the end-effector and a line that is centered and orthogonal to the front plane of the target cubby, $y_2=\phi_2(\x)=\|\x_c-\x\|$, where $\x, \x_c \in \mathbb{R}^3$ are the end-effector position, and the closest point on the line to the end-effector, where the line is orthogonal and centered with the front plane. The priority metric is  
\begin{align}
\label{eq:M_bw}
G_w (\x) = s(y_1)\Big(\big(\widebar{m} - \underline{m}\big)s(y_2) + \underline{m}\Big) \I.
\end{align}
where $s(y_1) = 1$ if $y_1<0.1$ and $s(y_1) = 0$, otherwise. $s(y_2)=0.5(\tanh(\alpha_m (y_2-r))+1)$, $\widebar{m}$, $\underline{m} $ are the upper and lower isotropic masses, respectively, and $\alpha_m $ defines the rate of transition between $0$ and $1$, while $r=0.15$ offsets the transition. Overall, the priority vanishes if the end-effector is either more than $0.1 \:  m$ away from the front plane (outside of the cubby) or within $0.15 \: m$ of the target cubby center line. The potential function is the same as defined in Eq.~\ref{eq:limit_potential}.

\textbf{Target Cubby Attraction:} An additional target attraction policy is used to help pull the end-effector inside a target cubby. It is the same as the end-effector attraction used in Layer 1, but with the priority metric defined as Eq.~\ref{eq:M_bw}. Overall, the priority remains zero until it enters a cylindrical region aligned with the target cubby. The heightening priority funnels motion into the cubby.

\textbf{Way-Point Attraction:} This term guides the arm to the cubby opening by attracting to a point $0.15m$ ahead of the front plane. The term generally matches the end-effector attraction term above, but with a different target and a switching function on the metric disabling it once within the column of the target cubby. Specifically, we make the following changes:
1) replace $x_t$ with $x_w$ in the task map, where $\x_w \in \mathbb{R}^3$ is the way point position; 2) define the switching function with the task map $y_2=\phi_2(\x)=\|\x_c-\x\|$ as described in the cubby extraction policy.
Note this term guides the arm but, as a geometric term, does not affect convergence to the target.
%A way point attraction policy is used to help pull the end-effector to the target cubby, and the way point is placed in front of the target cubby on the center line, with a distance of $0.15m$ away from the front plane. This policy is defined in the same way as the end-effector attraction, but with the following changes:
%1) replacing $x_t$ with $x_w$ in the task map, where $\x_w \in \mathbb{R}^3$ is the way point position, 2) a switching function defined with the task map $y_2=\phi_2(\x)=\|\x_c-\x\|$ as described in the cubby extraction policy, is used to disable this policy when the end-effector is within the column of the target cubby. 

\subsubsection{Layer 3}
This layer enables cubby collision avoidance.

\textbf{Cubby Collision:} The task map is $y = \phi_c(\x)$ which captures the minimum distance between a point on the arm and the cubby, where $\x$ in $\mathbb{R}^3$ is a designated {\em collision} point on the robot. 
Denoting the closest point on the cubby as $\x_c$, we use $y = \phi_c(\x) = \|\x_c - \x\|$ (ignoring dependencies of $\x_c$ on $\x$ for simplicity).
%If the point is inside of a cubby opening, the minimum distance is between the point and the closest wall, $y=\phi_c(\x)=\|\x-\x_w\|$, where $\x, \x_w \in \mathbb{R}^3$ are the end-effector location and its projection onto the closest wall, respectively. Otherwise, the minimum distance is between the point and the closest edge, $y=\phi(\x)=\|\x-\x_e\|$, where $\x, \x_e \in \mathbb{R}^3$ are the end-effector location and closest point on the closest edge. 
The metric is defined as a function of position, $G_b (y) = \frac{k_b}{y^2}$, where $k_b$ is a barrier gain. The acceleration-based potential gradient, $\partial_\q \psi_b(y) = M_b (y) \partial_\q \psi_{1,b}(y)$, uses $\psi_{1,b}(y) = \frac{\alpha_b}{y^8}$, where $\alpha_b$ is the barrier gain.

%Layer-wise construction of the geometries is critical for mitigating the complexity of design, enabling stage-wise behavioral design and even reuse of parts in different settings.

%The task objective is to continuously move the end effector to randomly selected targets located in different cubbies. To accomplish this task, we build the geometric fabric up layer-wise as follows. 

% See the longer version of this paper \cite{geometricFabricsForMotionArXiv2020} for a detailed description of each of these layers with specific functional forms.

\subsection{Objective and Damping}
The \textit{optimization potential} is an acceleration-based design as discussed in~\ref{subsec:acceleration_potential}, and it is designed for task space, $\y = \phi(\x) = \x - \x_t$, where $\x$, $\x_t \in \mathbb{R}^3$ are the current and target end-effector position in Euclidean space. The acceleration-based potential gradient, $\partial_\y \psi(\y) = M_\psi(\y) \partial_\y \psi_1(\y)$, uses $\psi_1(y)$ as defined in Eq.~\ref{eq:attractor_potential}, and the metric is defined as $G_{\psi}(\y) = \Big(\big(\widebar{m} - \underline{m}\big)s\big(\|\y\|\big) + \underline{m}\Big) \I,$ 
% \begin{align}
% \label{eq:M_psi}
% G_{\psi}(\y) = \Big(\big(\widebar{m} - \underline{m}\big)s\big(\|\y\|\big) + \underline{m}\Big) \I,
% \end{align}
where $s(\|\y\|)=0.5(\tanh(\alpha_m (\|\y\|-r))+1)$, $\widebar{m}$, $\underline{m}$ are the upper and lower isotropic masses, respectively, and $\alpha_m$ defines the rate of transition between $0$ and $1$, while $r$ offsets the transition. This design allows for small attraction when far away from the target while smoothly increasing priority as getting closer to the target.
\textit{Damping} is defined as in~\ref{subsec:damping_details}, where the gain $k$ in $\alpha_{boost}$ is defined as $k = -5\abs{\|\yd\|-\Lag_e^\mathrm{ex,d}}$, where $\|\yd\|$ is the current end-effector speed.

\subsection{Experiment Results}
\begin{figure}[!t]
  \centering
  \includegraphics[width=.98\linewidth]{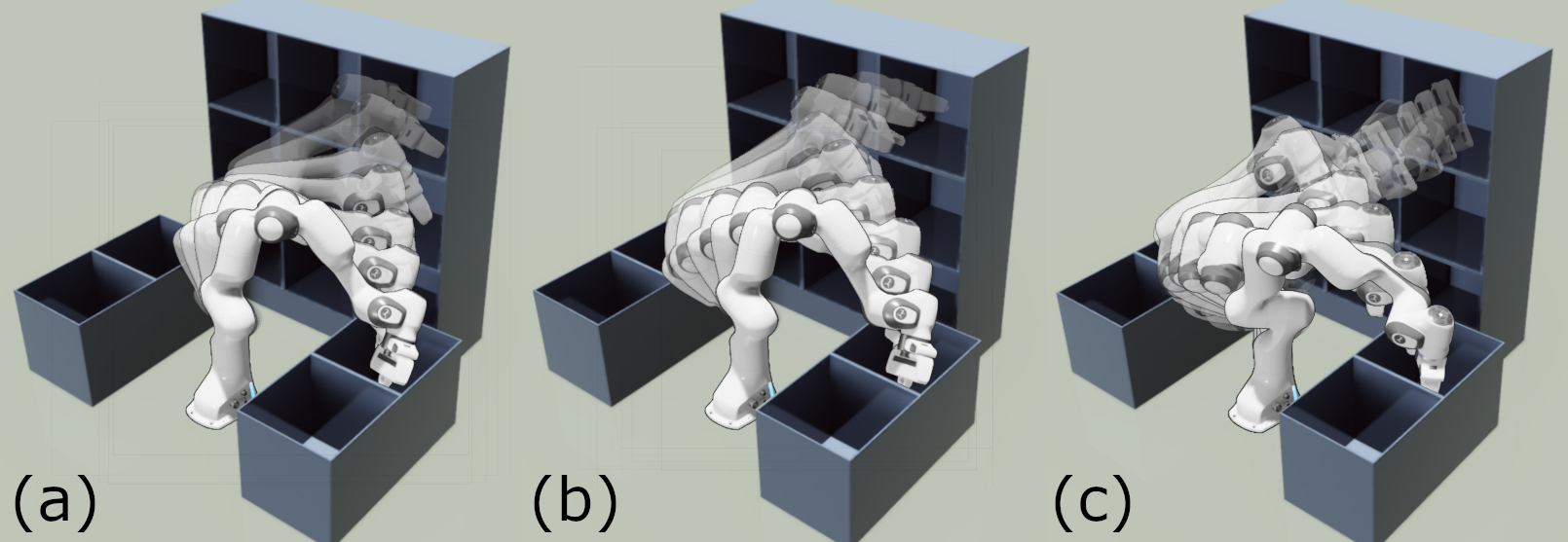}
  \vspace{-4mm}
  \caption{Robot navigation among cubbies when sequentially composing the geometric fabric with (a) Layer 1, (b) Layer 2, and (c) Layer 3.}
  \vspace{-4mm}
  \label{fig:layer_sweeps}
\end{figure}

As seen in Fig. \ref{fig:CubbyResults}, the robot intelligently navigates the cubbies. 
This global behavior was incrementally sequenced by adding layers of complexity to the underlying geometric fabric, a technique facilitated by geometric consistency and acceleration-based design. Fig. \ref{fig:layer_sweeps} shows the evolution of the robot trajectories for each of the three behavioral layers. Layer 1 depicts a more direct route to the target, ignoring the cubbies entirely. Layer 2 improves cubby navigation. Layer 3 enables complete collision avoidance. We dynamically change the desired end-effector goal during system execution, incurring rapid changes in system behavior to solve the desired task.

To demonstrate strong generalization, we hand tuned the system (analogous to training the system) with 6.7\% (6 of 90) of the available problems (navigating between a pair of cubbies), validated it on 90 problems with a 96.7\% success rate, and tested it on 90 problems with a success rate of 92.2\%, where the base was rotated by 30 degrees counterclockwise. Standard tuning protocol would iterate this cycle of tuning and testing until the system performed well on the entire problem distribution; this success rate is indicative of a generally fast tuning convergence. A second round of tuning easily solved the observed issues raising the success rate in both cases to 100\%.
We also tested the tuned policy on the 6 training problems with an array of perturbed variants,
%Despite using on only a handful of validation problems for tuning, the system generalizes well to all possible pairs of cubbies for navigation. Additionally, the system still performs well even if the robot base is perturbed, 
mimicking relocating the robot to the scene (such as with a mobile base). Perturbations to the base $(\Delta x,\Delta y,\Delta \theta)$ include combinations of $(- 0.1\: m, \pm 0.1 \: m, \pm \frac{\pi}{6} rad)$ and $(- 0.1\: m, \pm 0.1 \: m, 0 \: rad)$. The system behaved well on all variants.

\begin{figure}[!t]
  \centering
  \includegraphics[width=.8\linewidth]{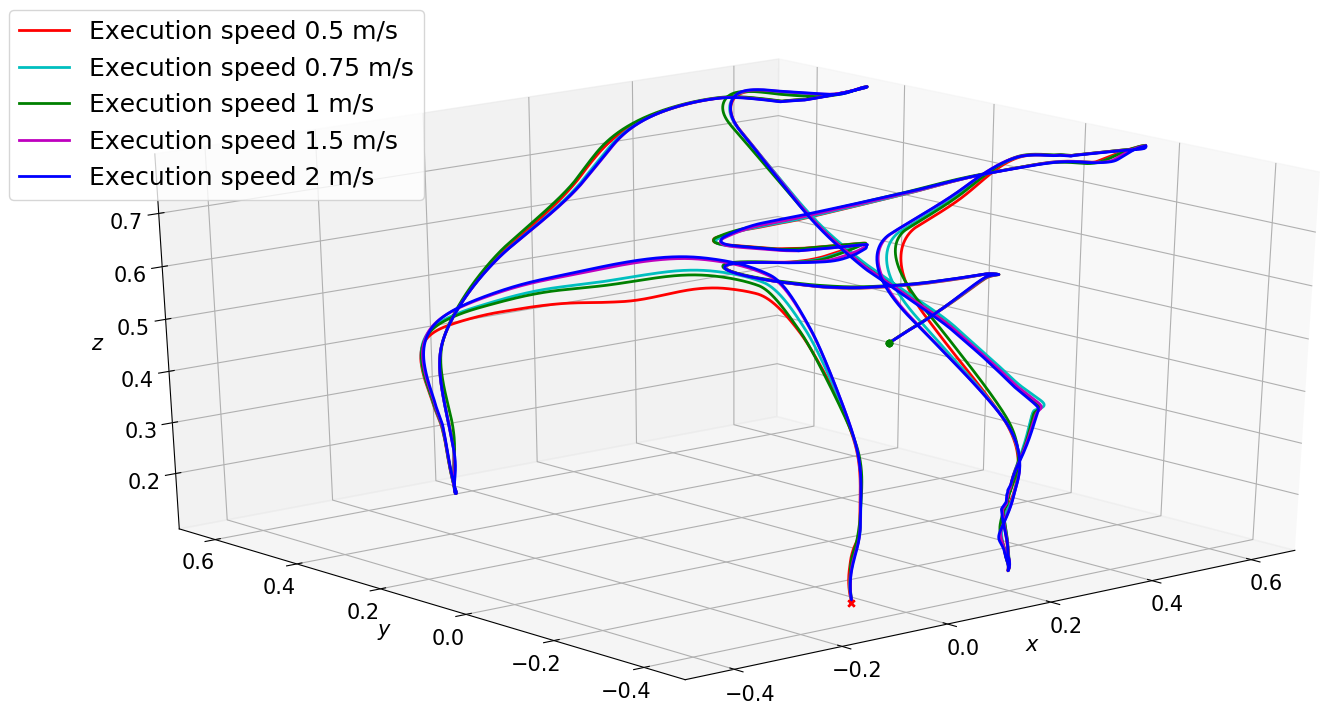}
  \vspace{-4mm}
  \caption{End-effector trajectories for five different execution speeds.}
  \label{fig:5speed}
  \vspace{-4mm}
\end{figure}

We additionally quantified path consistency across execution speeds. Five different execution speeds are tested for the same task with targets located in 7 different cubbies. The corresponding end-effector trajectories and speeds are shown in Fig~\ref{fig:5speed}. The minor differences among the paths are caused by the optimization potential pushing the system away from the unbiased nominal path defined by the geometry. Generally speaking, slower execution speeds subjects the system to greater influence from the potential, increasing the departure from path consistency. This is evident in Fig.~\ref{fig:5speed} where the fastest speeds (blue and purple curves) are nearly coincident and the slower speeds have greater changes to the paths taken. To facilitate path consistency, the optimization potential has a large gradient only when close to its minimum. 

We analyze path consistency in the configuration space by calculating the average arc length integral of each path against a cost defined as a function of the difference between this path and another one as $\mathbf{L} = \left(\sum_{t=0}^T \| \dot{q}_t \| \Delta t\right)^{-1}\sum_{t=0}^T c(q_t, Q) \| \dot{q}_t \| \Delta t$, where $\q$, $\qd \in \mathbb{R}^7$, and $c(q_t, Q)$ defines the minimum distance between a point at time $t$, namely $q_t$, and another path $Q$, and $\Delta_t = 0.001\:s$. Based on this definition, we would expect the average arc length integrals for each path to be zero if the path matches each other perfectly. The average arc length integral of each path with respect to the other ones in the configuration space is shown the Table~\ref{table:config_path_diff}, and from which we can observe that the arc length integral increases as the difference in the execution speed increases.

\begin{table}[!htb]
	\caption{arc length integrals in the configuration space.}
	\label{table:config_path_diff}
	\begin{center}
	    \vspace{-8mm}
		\begin{tabular}{| c | c | c | c | c | c |}
			\hline 
			Speed (m/s)
			& 0.5 & 0.75 & 1.0 & 1.5 & 2.0 \\
			\hline 
			0.5 & 0.0 & 0.0473 & 0.0550 & 0.0992 & 0.1193 \\
			\hline 
			0.75 & 0.0473 & 0.0 & 0.0444 & 0.0619 & 0.0868 \\
			\hline 
			1.0 & 0.0550 & 0.0444 & 0.0 & 0.0895 & 0.1110 \\
			\hline 
            1.5 & 0.0992 & 0.0619 & 0.0895 & 0.0 & 0.0268 \\
			\hline 
			2.0 & 0.1193 & 0.0868 & 0.1110 & 0.0268 & 0.0 \\
			\hline 
		\end{tabular}
	\end{center}
	\vspace{-4mm}
\end{table}

\fi % End full version only
\section{CONCLUSIONS}

Geometric fabrics are RMPs whose acceleration policies are HD2 geometries and whose priority metrics derive from Finsler energies.
They constitute compact encoding of behavior simple enough to shape by hand.
They rival the broadest class of RMPs \cite{ratliff2018rmps} in expressivity and add a geometric consistency enabling simple layer-wise design and speed regulation. We document concrete design tools and layering procedures effective in realistic settings.
Interestingly, the strong generalization observed in our experiments suggests geometric fabrics may additionally represent a well-informed and flexible inductive bias for policy learning.

\clearpage
%% Use plainnat to work nicely with natbib. 
\bibliographystyle{plainnat}
\bibliography{refs.bib}
\clearpage

\begin{appendices}
\title{Geometric Fabrics for the Acceleration-based Design of Robotic Motion: Appendices}
\date{}
\maketitle

\section{Controlled Particle Experiments} \label{sec:ControlledParticleExperiments}
These experiments validate the theoretical findings and generate insight into the behavioral differences between geometric fabrics and Lagrangian fabrics, and usage of speed control. A set of 14 particles at rest to the right of a circular obstacle are attracted to a point to the left (small square). Behavior is integrated for 15 seconds using a fourth order Runge-Kutta method.

% \nathan{Is the information in the Point Attraction and Circular Object Repulsion consistent with the above. Perhaps remove or merge up there? I think this entire section can effectively just be the content in the paragraph above and the experiments section (C).}

\subsection{Point Attraction}
\label{subsec:point_attraction}
The behavior for pulling the end-effector towards a target location is constructed as follows. The task map is $\x = \phi(\q) = \q - \q_d$, where $\q$, $\q_d \in \mathbb{R}^2$ are the current and desired particle position in Euclidean space. The metric is designed as
\begin{align}
\label{eq:potential_priority}
\G_\psi(\x) = (\widebar{m} - \underline{m}) e^{-(\alpha_m \|\x\|) ^ 2} I + \underline{m} I.
\end{align}
where $\widebar{m}$, $\underline{m} \in \mathbb{R}^+$ are the upper and lower isotropic masses, respectively, and $\alpha_m \in \mathbb{R}^+$ controls the width of the radial basis function. For the following experiments, $\widebar{m} = 2$, $\underline{m}=0.2$, and $\alpha_m = 0.75$. The acceleration-based potential gradient, $\partial_\q \psi(\x) = M_\psi(\x) \partial_\q \psi_1(\x)$, is designed with
\begin{align}
\label{eq:attractor_potential}
\psi_1(\x) = k \left( \|\x\| + \frac{1}{\alpha_\psi}\log(1 + e^{-2\alpha_\psi \|\x\|} \right)
\end{align}
where $k \in \mathbb{R}^+$ controls the overall gradient strength, $\alpha_\psi \in \mathbb{R}^+$ controls the transition rate of $\partial_\q \psi_1(\x)$ from a positive constant to 0. For these experiments, $k = 10$, $\alpha_\psi = 10$.

For Lagrangian Fabrics, the policy is created from the Lagrangian, $\Lag = \xd^T \G_\psi(\x) \xd - 
\psi(\x)$. For geometric fabrics, the policy is constructed as $\xdd = -\partial_\x \psi_1(\x)$ and weighted by $\M_\psi(\x)$ from the Finsler energy, $\Lag = \xd^T \G_\psi(\x) \xd$.

\subsection{Circular Object Repulsion}
\label{subsec:circular_object_repulsion}
Collision avoidance with respect to a circular object is constructed as follows. The task map is $x = \phi(\q) = \frac{\|\q - \q_o\|}{r} - 1$, where $\q_o$ is the origin of the circle and $r$ is its radius. Two different metrics are designed to prioritize this behavior. The first is just a function of position, $G_b (x) = \frac{k_b}{x^2}$, while the second is a function of both position and velocity, $G_b (x, \dot{x}) = s(\dot{x}) \frac{k_b}{x^2}$. Moreover, $k_b \in \mathbb{R}^+$ is a barrier gain and $s(\dot{x})$ is a velocity-based switching function. Specifically, $s(\dot{x}) = 1$ if $\dot{x} < 0$ and $s(\dot{x}) = 0$, otherwise. For these experiments, $k_b = 20$.

The acceleration-based potential gradient, $\partial_\q \psi_b(x) = M_b (x) \partial_\q \psi_{1,b}(x)$, is designed with $\psi_{1,b}(x) = \frac{\alpha_b}{2 x^8}$, where $\alpha_b \in \mathbb{R}^+$ is the barrier gain and $\alpha_b = 1$ for these experiments. 

For Lagrangian fabrics, the policy is created from either $\Lag = G_b(x) \dot{x}^2 - 
\psi_b(x)$ or  $\Lag = G_b(x, \dot{x}) \dot{x}^2 - 
\psi_b(x)$. For geometric fabrics, the policy is $
\ddot{x} = - s(\dot{x}) \dot{x}^2 \partial_x \psi_{1,b}(x)$. This policy is weighted by $\M_b(x)$ or $\M_b(x, \dot{x})$.

\subsection{Experiments}
\begin{figure}[!t]
  \centering
  \includegraphics[width=.8\linewidth]{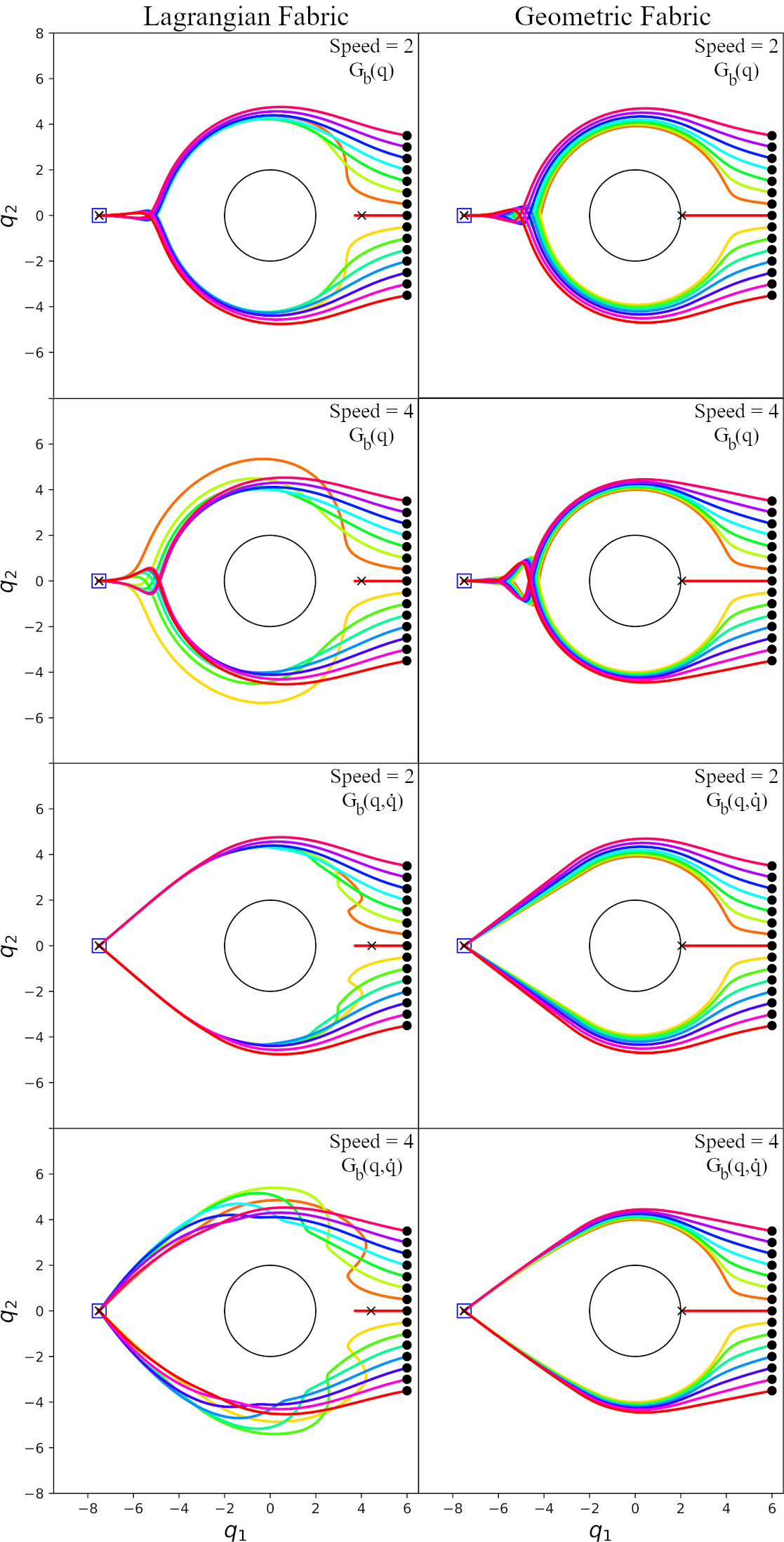}
  \vspace{-4mm}
  \caption{Particle behavior for Lagrangian and geometric fabrics with different metric designs.}
%   \vspace{-4mm}
  \label{fig:gf_vs_lds_particles}
\end{figure}

Speed control is used with the execution energy designed as $\Lag_{ex} = \frac{1}{v_d} \qd^T \qd$, where $v_d \in \mathbb{R}^+$ is the desired Euclidean speed. For both geometric and Lagrangian fabrics, behavior is simulated for $v_d=2,4$, for cases using $G_b(x)$ and $G_b(x, \dot{x})$ (see Fig. \ref{fig:gf_vs_lds_particles}).

All particles successfully circumnavigate the object and reach the the target position. However, the Lagrangian fabrics produce more inconsistent behavior across the speed levels. This is pronounced when using $G_b(x, \dot{x})$ since the velocity gate does not modulate the entire obstacle avoidance policy for Lagrangian fabrics. Instead, the mass of the obstacle avoidance policy vanishes, while components of its force remain. The effect of these forces are amplified when traveling at a higher velocity, producing the ``launching'' artifacts. In contrast, geometric fabrics produce more consistent paths across speed levels without any launching artifacts. Finally, the velocity gate facilitates straight-line motion to the desired location once the obstacle is bypassed. Without this gate the system possesses large mass, impeding motion to the desired location.

\begin{figure}[!t]
  \centering
  \includegraphics[width=1.\linewidth]{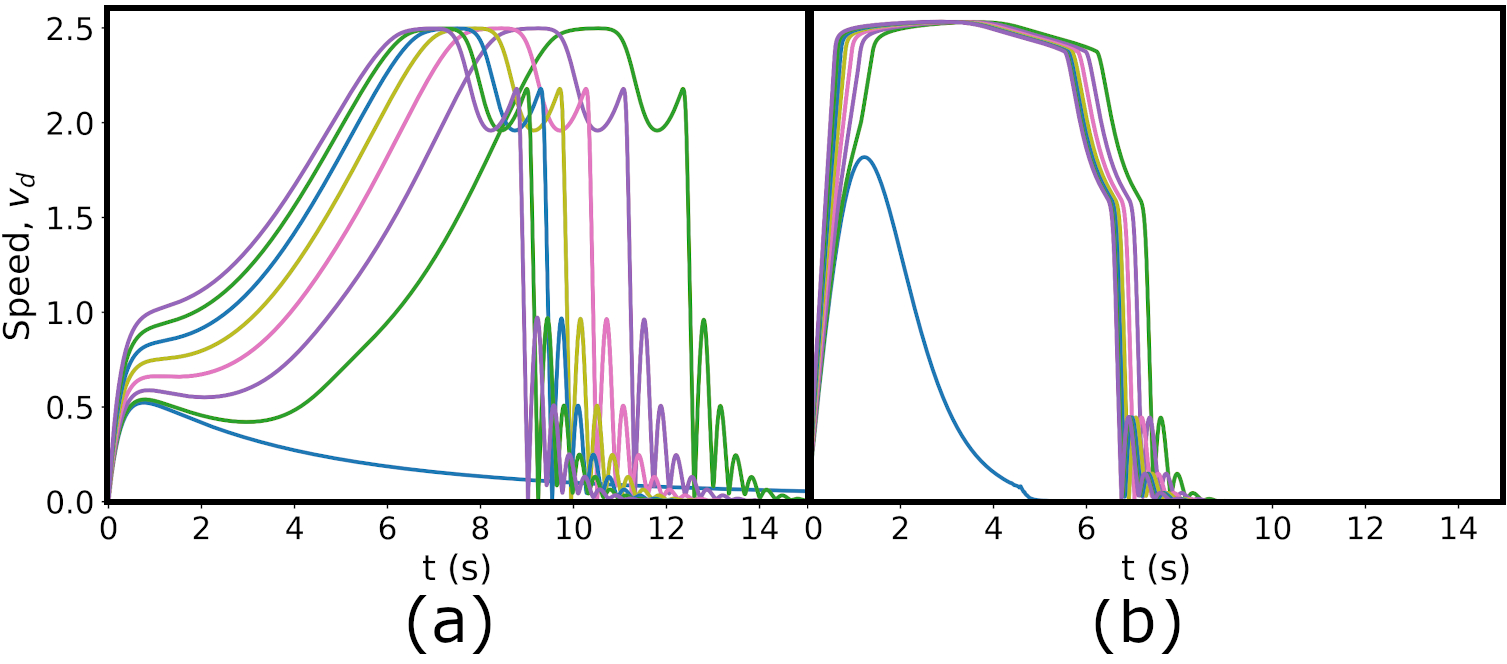}
  \vspace{-6mm}
  \caption{Speed profiles over time when optimizing over a geometric fabric with (a) basic damping and (b) speed control.}
  \vspace{-4mm}
  \label{fig:speed_vs_basic}
\end{figure}

Using the same geometric fabric (with obstacle velocity gate) and optimization potential from the previous section, the system is evaluated using: 1) speed control with $v_d=2.5$, and 2) basic damping with $\beta = 4$. The resulting time-speed traces are shown in Fig. \ref{fig:speed_vs_basic}. While both strategies still optimize, basic damping leads to a rather chaotic speed profile over time. In contrast, speed control achieves the desired speed in approximately 1 second and maintains it before damping to the final convergence point.

\end{appendices}

\end{document}